%% file: main.tex
\documentclass[moor]{informs_arxiv}              

\usepackage{natbib}
\NatBibNumeric
\def\bibfont{\small}%
\bibpunct[, ]{[}{]}{,}{n}{}{,}%

\usepackage[colorlinks=true,breaklinks=true,bookmarks=true,urlcolor=blue,
citecolor=blue,linkcolor=blue,bookmarksopen=false,draft=false]{hyperref}

\def\EMAIL#1{\href{mailto:#1}{#1}}

\TheoremsNumberedThrough     

\EquationsNumberedThrough    

\input{packages}

\input{notations}

\newif\ifsup
\supfalse\suptrue		

\begin{document}
	\RUNAUTHOR{Verma et al.}

	\RUNTITLE{Censored Semi-Bandits for Resource Allocation}

	\TITLE{Censored Semi-Bandits for Resource Allocation}

	\ARTICLEAUTHORS{
		\AUTHOR{Arun Verma}
		\AFF{Indian Institute of Technology Bombay, \EMAIL{v.arun@iitb.ac.in} }
		\AUTHOR{Manjesh K. Hanawal}
		\AFF{Indian Institute of Technology Bombay, \EMAIL{mhanawal@iitb.ac.in} }
		\AUTHOR{Arun Rajkumar}
		\AFF{Indian Institute of Technology Madras, \EMAIL{arunr@cse.iitm.ac.in} }
		\AUTHOR{Raman Sankaran}
		\AFF{LinkedIn India, \EMAIL{rsankara@linkedin.com} }
	}

	\ABSTRACT{
		We consider the problem of sequentially allocating resources in a \emph{censored semi-bandits} setup, where the learner allocates resources at each step to the arms and observes loss. The loss depends on two hidden parameters, one specific to the arm but independent of the resource allocation, and the other depends on the allocated resource. More specifically, the loss equals zero for an arm if the resource allocated to it exceeds a constant (but unknown) arm dependent threshold. The goal is to learn a resource allocation that minimizes the expected loss. The problem is challenging because the loss distribution and threshold value of each arm are unknown. We study this setting by establishing its `equivalence' to Multiple-Play Multi-Armed Bandits (MP-MAB) and Combinatorial Semi-Bandits. Exploiting these equivalences, we derive optimal algorithms for our problem setting using known algorithms for MP-MAB and Combinatorial Semi-Bandits. The experiments on synthetically generated data validate the performance guarantees of the proposed algorithms.
	}


	\maketitle

	\section{Introduction}
	\label{sec:introduction}
	\input{introduction}

	\section{Problem Setting}		
	\label{sec:problemSetting}
	\input{problem_setting}

	\section{CSB problems with Known $T$ and $\epsilon$}
	\label{sec:known}
	\input{known}

	\section{Anytime and parameter agnostic algorithms}
	\label{sec:unknown}
	\input{unknown}

	\section{Experiments}
	\label{sec:experiments}
	\input{experiment}

	\section{Conclusion and Future Extensions}
	\label{sec:conclusion}
	\input{conclusion}

	\begin{APPENDICES}
		\section{Appendix}
		\label{sec:appendix}
		\input{appendix}

		\section{CSB for Stochastic Network Utility Maximization}
		\label{sec:num}
		\input{num}

	\end{APPENDICES}

	\bibliographystyle{informs} 
	\bibliography{ref} 
	
\end{document}

%% file: packages.tex
\usepackage[markup=underlined]{changes}

\usepackage[utf8]{inputenc}   
\usepackage[T1]{fontenc}    	
\usepackage{url}              		 
\usepackage{booktabs}       	
\usepackage{amsfonts}       	
\usepackage{nicefrac}       	  
\usepackage{microtype}      	

\usepackage{graphicx}
\usepackage{color}
\usepackage{rotating}
\usepackage{tabularx}
\usepackage{pdflscape}
\usepackage{amsmath,amssymb}
\usepackage{algorithm,algpseudocode}
\usepackage{bbm,dsfont}
\usepackage{subcaption}
\usepackage{wrapfig}
\usepackage{cancel}
\usepackage{enumerate, cases}
\usepackage{thmtools,thm-restate}
\usepackage{mathtools}

\usepackage[none]{hyphenat}
\usepackage{multicol,multirow,diagbox}
\usepackage{makecell}
\usepackage{setspace}
\usepackage{pifont}
\usepackage{array}
\usepackage{enumitem}
\usepackage{fullpage}
\usepackage{anyfontsize}

\allowdisplaybreaks

\usepackage[capitalize]{cleveref}

%% file: notations.tex
\newcolumntype{C}[1]{>{\centering}m{#1}}
\newcommand{\Regret}{\mathcal{R}}
\newcommand{\A}{\cA_{Q}}
\newcommand{\PCSB}{\mathcal{P}} 

\newcommand{\PMP}{\cP_{\text{MP}}}
\newcommand{\PCoSB}{\mathcal{P}_{\text{CoSB}}}
\newcommand{\bmu}{\boldsymbol{\mu}}
\newcommand{\btheta}{\boldsymbol{\theta}}
\newcommand{\ba}{\boldsymbol{a}}

\newcommand{\EE}[1]{\bE\left[#1\right]}

\newcommand{\Prob}[1]{\bP\left\{#1\right\}}

\newcommand{\R}{\bR}
\newcommand{\N}{\bN}

\newcommand{\one}[1]{\mathds{1}_{\left\{#1\right\}}}

\renewcommand{\phi}{\varphi}
\renewcommand{\epsilon}{\varepsilon}

\newcommand{\el}{\end{flushleft}}
\newcommand{\bl}{\begin{flushleft}}
\newcommand{\floor}[1]{\left\lfloor #1 \right\rfloor}
\newcommand{\ceil}[1]{\left\lceil #1 \right\rceil}

\newcommand{\bE}{\mathbb{E}}

\newcommand{\bN}{\mathbb{N}}

\newcommand{\bP}{\mathbb{P}}

\newcommand{\bR}{\mathbb{R}}

\newcommand{\cA}{\mathcal{A}}

\newcommand{\cM}{\mathcal{M}}

\newcommand{\cP}{\mathcal{P}}



%% file: introduction.tex

In the classical multi-armed bandit setup, the assumption is that the learner always observes a loss/reward sample as feedback by playing arms (or actions). In many applications, the learner first needs to assign the resources to the arms, and depending on the allocated resource, the loss may or may not be observed from the selected arms. When the loss is not observed, we say that `feedback is censored' and refer to the case as `censored feedback.' Sequential allocation problems with censored feedback have received significant interest in recent times as censoring occurs naturally in several applications. Some of the examples are:

\noindent
{\bf Example 1:} (Policing and poaching control) In opportunistic crime/poaching control, the goal is to minimize total crimes in some regions using available manpower. For this, the police may spread its manpower (resource allocation) across the regions (arms) for patrolling \citep{AOR14_adler2014location, NSE10_curtin2010determining, AAMAS18_gholami2018adversary, AAMAS16_nguyen2016capture, IJCAI17_rosenfeld2017security}. A thief/poacher intending to commit a crime may abstain from committing a crime if the patrol is heavy, otherwise, continue with his plan. Thus, the censoring of feedback occurs when the thief/poacher intending to commit a crime abstains due to fear of getting caught.

\noindent
{\bf Example 2:} (Auctions) In the auction of multiple items (arms), a bidder with fixed budge decides the amount to bid (resource) for each item \citep{MS19_balseiro2019learning, NIPS17_baltaoglu2017online, Allerton11_gummadi2011optimal, ICML14_mohri2014learning, COLT16_weed2016online}.  The bidder gets to see the item's actual worth (feedback) only if she wins; otherwise, they do not see it (censored feedback). Here, the winning of an item depends on the bidding amount.

\noindent
{\bf Example 3:} (Network Utility Maximization) Power is a scarce resource in wireless networks. In multi-channel communication, nodes need to split the power across the channels to maximize their sum-rate \citep{ETT1997_ChargingAndRateControl, INFOCOM20_verma2020stochastic}. Unless a node transmits with enough power level on a channel, its transmission always fails and succeeds with a certain probability when transmitted power is above a certain threshold. Thus, the recipient gets to observe the channel quality only when enough power is given to the nodes; otherwise, it is censored.

Censoring of feedback also occurs in the problem of supplier selection  \citep{NIPS16_abernethy2016threshold}, budget allocation  \citep{UAI12_amin2012budget,ALT18_dagan18a,UAI14_lattimore2014optimal,NIPS15_lattimore2015linear}, and several others. In all these applications, unless enough resource is applied to an arm, the feedback from arms gets censored. The challenge in these problems is how to learn the quality of all the arms by appropriately allocating the available resource and then optimally allocating resources to minimize the total loss incurred.

Classical approaches to this problem are to learn from historical data \citep{AOR14_adler2014location, NSE10_curtin2010determining, IJCAI17_rosenfeld2017security, AAMAS16_zhang2016using}.  Game-theoretic approaches have also been considered \citep{AAMAS18_gholami2018adversary, AAMAS16_nguyen2016capture, IJCAI18_sinha2018stackelberg}, where the user (buyer, criminal, etc.) knows the history of allocations and responds strategically. While the classical approach of learning from historical data fails to capture the problem's sequential nature, the game-theoretic approach is agnostic to the user (buyer, criminal, etc.) behavioral modeling. In this work, we balance these two approaches by proposing a simple yet novel threshold-based user behavioral model, which we term as \emph{Censored Semi-Bandits} (CSB)\footnote{This paper is an extended version of \cite{NeurIPS19_verma2019censored} published in Neural Information Processing Systems (NeurIPS 2019).}.
Under the CSB model, the loss incurred from each arm follows a generative structure. The learner has access to a fixed amount of resources in each round which can be allocated to the arms. Each arm has an associated threshold that decides whether the learner observes reward from that arm: if the arm receives resources below a threshold, the learner observes a loss from that arm; otherwise, no loss value is observed. The threshold captures behaviors of the arms. For example, in the crime control problem, the threat perception of a thief/poacher being caught in an area in the presence of patrolling determines the threshold level in that area.

In the first variation of our proposed behavioral models, we assume the threshold (user behavioral) is uniform across arms (set of options). We establish that this setup (with known threshold) is `equivalent' to Multiple-Play Multi-Armed Bandits (MP-MAB), where a fixed number of arms is played in each round. We also study the more general variation, where the threshold is arm dependent. We establish that this setup (with known threshold) is equivalent to Combinatorial Semi-Bandits, where a subset of arms to be played is decided by solving a combinatorial $0$-$1$ knapsack problem. Formally, we tackle the sequential nature of the resource allocation problem by establishing its equivalence to the MP-MAB and Combinatorial Semi-Bandits framework. By exploiting this equivalence for our proposed threshold-based behavioral model, we develop novel resource allocation algorithms by adapting existing algorithms and providing optimal regret guarantees. More precisely, we make the following contributions in this paper that substantially extend the algorithms and results given in \cite{NeurIPS19_verma2019censored}:
\begin{itemize} 
    \item In \cref{sec:known}, we improve the state-of-the-art horizon dependent algorithms \citep{NeurIPS19_verma2019censored} for estimating the thresholds and mean losses with arms. The new algorithms are simpler and have better empirical performance as these algorithms collect the loss information during the threshold estimation. 
    
    \item We develop a novel sequential resource allocation algorithm to the CSB problem with multiple thresholds (the number of thresholds can be smaller than the number of arms). We prove that the regret bound of the algorithm is sub-linear and depends on the number of unique thresholds. We also show empirically that the proposed algorithms have better regret performance.
    
    \item The algorithms in \cref{sec:known} are horizon $(T)$ dependent and requires the minimum mean loss $(\epsilon)$ and an accuracy tolerance $(\delta)$ that decides the stopping criteria for the threshold estimation method as input.  In \cref{sec:unknown}, we develop anytime algorithms that do not need $T$, $\epsilon$, and $\delta$ as input. The anytime algorithms use a linear search based method to estimate the thresholds, which is different from the binary search based method used in the horizon dependent algorithms.

    \item We extend the CSB setup to reward maximization setting by discussing the stochastic Network Utility Maximization problem (NUM). In the reward setting, anytime algorithms developed for the loss setting cannot be applied directly. We give algorithms that work with the known value of the time horizon. The details are given in \cref{sec:num}.
\end{itemize}

\subsection{Related Work}
The problem of resource allocation in many areas has received significant interest in recent times. Several directions have been considered in resource allocation problems to tackle crime \citep{NSE10_curtin2010determining, AAMAS18_gholami2018adversary, AAMAS16_nguyen2016capture}, some of which learn from historical data while others are game-theoretic. \cite{NSE10_curtin2010determining} employ a static maximum coverage strategy for spatial police allocation while  \cite{AAMAS18_gholami2018adversary} and \cite{AAMAS16_nguyen2016capture} study game-theoretic and adversarial perpetrator strategies. We, on the other hand, restrict ourselves to a stochastic setting. The work in  \citep{AOR14_adler2014location, IJCAI17_rosenfeld2017security} look at traffic police resource deployment and consider the optimization aspects of the problem using real-time traffic, etc., which differs from the main focus of our work. \cite{AAMAS15_zhang2015keeping} investigates dynamic resource allocation in the context of police patrolling and poaching for opportunistic criminals. Here, they attempt to learn a model of criminals using a dynamic Bayesian network. Our approach proposes simpler and realistic modeling of perpetrators, where we exploit the underlying structure effectively and efficiently.

We pose our problem in the exploration-exploitation paradigm, which involves solving the MP-MAB and combinatorial 0-1 knapsack problem. It is different from the bandits with Knapsacks setting studied in \cite{JACM18_badanidiyuru2018bandits}, where resources get consumed in every round. The work of \cite{NIPS16_abernethy2016threshold}, \cite{Arxiv20_bengs2020multi}, and \cite{ICML18_jain2018firing} are similar to us in the sense that they are also threshold-based settings. However, the thresholding we employ naturally fits our problem and significantly differs from theirs. Specifically, their thresholding is either on a sample generated from an underlying distribution \citep{NIPS16_abernethy2016threshold, ICML18_jain2018firing} or chosen by the learner \cite{Arxiv20_bengs2020multi} in each round. In contrast, we work in a Bernoulli setting where the thresholding is based on the allocation. Resource allocation with semi-bandits feedback   \citep{ALT18_dagan18a, ALT20_fontaine2020adaptive, UAI14_lattimore2014optimal, NIPS15_lattimore2015linear} is also a related but less general setup where the reward is based only on allocation and a hidden threshold. Our setting requires an additional unknown parameter for each arm, a `mean loss,' which also affects the reward. 
When the learner observes no loss in the CSB setup, it is difficult to say whether it is an actual loss or a censored loss due to enough resource allocation. This dilemma leads to the learner's inability to infer loss from observed feedback when enough resources are allocated to arms. The extreme forms of such problems are studied in \citep{AISTATS19_verma2019online, ACML20_verma2020thompson, NeurIPS20_verma2020online}, where the learner can not infer the loss/ reward from the observed feedback.

Resource allocation problems in the combinatorial setting have been explored in \citep{JCSS12_cesa2012combinatorial, NIPS16_chen2016combinatorial, ICML13_chen2013combinatorial, NIPS15_combes2015combinatorial, NeurIPS20_perrault2020statistical,NIPS14_rajkumar2014online, ICML18_wang2018thompson}. Even though these are not related to our setting directly, we derive explicit connections to the sub-problem of our algorithms to the setup of \cite{ICML15_komiyama2015optimal} and \cite{NeurIPS20_perrault2020statistical}.

%% file: problem_setting.tex

\begin{figure}[!ht]
	\centering
	\includegraphics[width=0.4\linewidth]{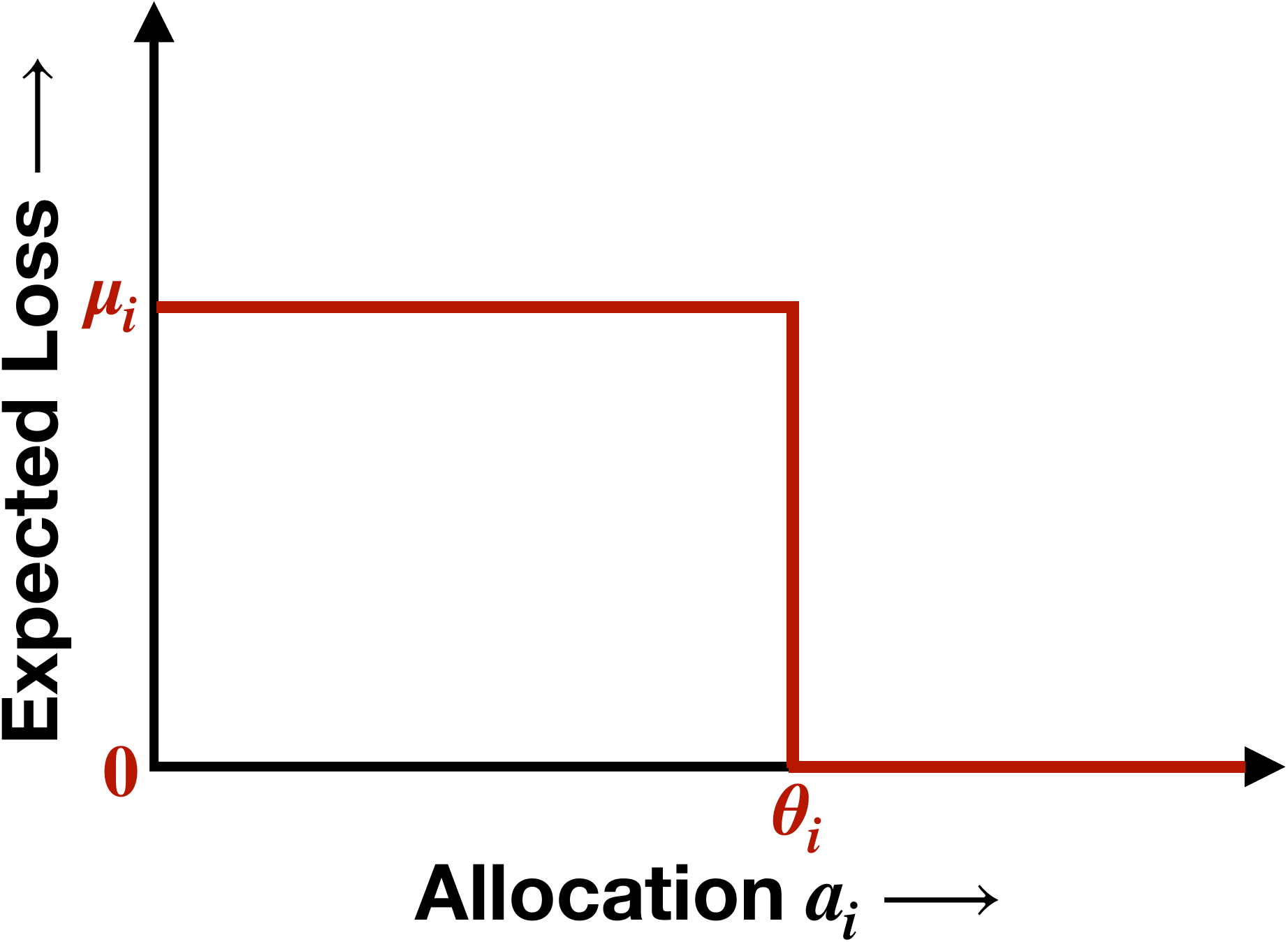}
	\caption{Relationship between allocated resources $(a_i)$ and mean loss $(\mu_i)$ of an arm $i$.}
	\label{fig:ThresholdFunction}
\end{figure}

We consider a sequential learning problem where $K$ denotes the number of arms, and $Q$ denotes the amount of divisible resources. The loss at arm $i \in [K]$ where $[K] := \{1, 2, \ldots, K\}$, is Bernoulli distributed with mean $\mu_i \in [0, 1]$ and independent and identically distributed (IID), whose realization in the $t^{th}$ round is denoted by $X_{t, i}$. Each arm may be assigned a fraction of resources, which determines the feedback observed and the loss incurred from that arm. Formally, denoting the resources allocated to the arms by $\ba:=\{a_i: i\in [K]\ |\ a_i \in [0, Q]\}$, the loss incurred equals the realization of the arm $X_{t, i}$ if $a_i < \theta_i$, where $\theta_i \in [0, Q]$ is fixed but unknown threshold\footnote{One could consider a smooth function instead of a step function, but the analysis is more involved, and our results need not generalize straightforwardly.}. When $a_i \ge \theta_i$, which corresponds to the scenario when the allocated resources are more than the threshold, we do not observe $X_{t, i}$, and hence the loss equals $0$. \cref{fig:ThresholdFunction} depicts the relationship between allocated resources and mean loss of an arm. For each $i \in [K]$, $\theta_i$ denotes the threshold associated with arm $i$ and is such that a loss is incurred at arm $i$ only if $a_i< \theta_i$. An allocation vector $\ba$ is said to be feasible if $\sum_{i \in [K]} a_i \leq Q$ and set of all feasible allocations is denoted as $\A$. The goal is to find a feasible resource allocation that results in a maximum reduction in the total mean loss. 

In the CSB setup, the learner allocates resources to multiple arms. However, loss from the arms may not be observed depending on the amount of resources allocated to them. We thus have a version of the partial monitoring system \citep{MOR14_bartok2014partial,ICML12_bartok2012partial,MOR06_cesa2006regret} with semi-bandit feedback. The vectors $\btheta =\{\theta_i\}_{ i\in [K]}$ and $\bmu =\{\mu_j\}_{i \in [K]}$ are unknown and identify an instance of CSB problem, which we denote henceforth using $P=(\bmu, \btheta, Q) \in [0,1]^{K}\times \R_+^K \times \R_+$. The collection of all CSB instances is denoted as $\PCSB$. For simplicity of discussion, we assume that means are ordered as $\mu_1 \ge  \mu_2 \ge \ldots \ge \mu_K$ and for any integer $M$, refer to the first $M$ arms in the order as the top-$M$ arms. Of course, the algorithm is not aware of this order. For instance $P \in \PCSB$ with known $\bmu$, $\btheta$, and $Q$, the optimal allocation can be computed by solving the following $0$-$1$ knapsack problem:
\begin{equation*}
	\ba^\star  \in \argmin _{\ba \in \A} \sum_{i=1}^K \mu_i \one{a_i< \theta_i}.
\end{equation*}
Here, $\A = \{\ba \in [0, Q]^K | \sum_{i} a_i \le Q\}$ denotes the set of all feasible resource allocations.  Since $\bmu$ and $\btheta$ are unknown, we estimate them in an online fashion using the observations made in each round. 
The interaction between the environment and a learner is given in \ref{alg:protocol}.
\begin{algorithm}[!ht]
	\renewcommand{\thealgorithm}{Algorithm 1}
	\floatname{algorithm}{}
	\caption{CSB Problem with instance $(\bmu, \btheta, Q)$}
	\label{alg:protocol}
	In round $t$: 
	\begin{enumerate}
		\setlength{\itemsep}{0pt}
		\item \textbf{Environment} generates a vector $\boldsymbol{X_t} = (X_{t,1}, X_{t,2},\ldots, X_{t,K}) \in \{0,1\}^K$, where $\EE{X_{t,i}}=\mu_i$ and the sequence $(X_{t,i})_{t\geq 1}$ is IID for all $i\in [K]$   
		\item \textbf{Learner} picks an resource allocation vector $\ba_t \in \A$
		\item \textbf{Feedback and Loss:} The learner observes a random feedback $\boldsymbol{Y_t}=\{Y_{t,i}: i\in [K]\}$, where $Y_{t,i}=X_{t,i}\one{a_{t,i}<\theta_i}$ and incurs loss $\sum_{i \in [K]}Y_{t,i}$
	\end{enumerate}
\end{algorithm}

We aim to design optimal strategies that accumulate minimum mean loss and measure its performance by comparing its mean cumulative loss with that of an Oracle that makes the optimal resource allocation in each round. Specifically, we define regret for $T$ rounds as
\begin{equation*}
\EE{\Regret_T} =  \sum_{t=1}^T\sum_{i=1}^K \mu_i \one{a_{t,i}< \theta_i} -   \sum_{t=1}^T\sum_{i=1}^K \mu_i \one{a^\star_i< \theta_i}.
\end{equation*}

Note that minimizing the mean cumulative loss of a policy is the same as minimizing the policy's regret. Our goal is to learn a policy that gives sub-linear expected regret, i.e., $\EE{\Regret_T}/T \rightarrow 0$ as $T \rightarrow \infty$. It implies that a leaner collects almost as much reward in the long run as an oracle who knows the best action from the beginning.

\subsection{Allocation Equivalent}
Next, we define when a pair of threshold vectors for the given loss vector and resources to be `equivalent.'
\begin{definition}[Allocation Equivalent]
	For fixed loss vector $\bmu$ and resources $Q$, two threshold vectors $\btheta$ and $\hat{\btheta}$ are {\em allocation equivalent} if the following holds:
	\begin{equation*}
	\min_{\ba \in \A} \sum_{i=1}^K\mu_i \one{a_i\ge \theta_i}  = 
	\min_{\ba \in \A} \sum_{i=1}^K\mu_i \one{a_i\ge \hat{\theta}_i}.
	\end{equation*}
\end{definition}
In simple words, we say that two threshold vectors $\btheta$ and $\hat{\btheta}$ are allocation equivalent if the minimum mean loss in instances $(\bmu,\btheta, Q)$ and $(\bmu, \hat{\btheta}, Q)$ are the same for fixed loss vector $\bmu$ and resource $Q$. Such equivalence allows us to estimate the threshold vector within some tolerance.

For ease of exposition and to bring out the algorithmic ideas clearly, in \cref{sec:known}, we start with a setting where we assume that time horizon ($T$) is known and mean rewards are larger than some known $\epsilon>0$, i.e., $\mu_i\geq\epsilon$ for all $i\in[K]$. This setup will aid in connecting our problem with the Multi-player bandits. In \cref{sec:unknown}, we relax these assumptions and develop anytime algorithms that do not need to know $\epsilon$. The algorithms in \cref{sec:known}, are based on binary search methods, while that in \cref{sec:unknown}, are based on linear search methods.

%% file: known.tex

This section introduces the algorithms for solving the CSB problem, where the time horizon ($T$) and the lower bound on the mean losses ($\epsilon$) are known. With this information, we can estimate the allocation equivalent using a binary search based method. Once allocation equivalent is known, the mean losses are estimated, and accordingly, resources are allocated among the arms. We first study a simple case where all arms have the same threshold and then study the more general case where all arms may not have the same threshold.

\subsection{Arms with Same Threshold}
\input{known_same}

\subsection{Arms with Multiple Threshold}
\input{known_multiple}

%% file: known_same.tex

We first focus on the simple case, where the threshold of all arms are the same, i.e., $\theta_i=\theta_s$ for all $i \in [K]$ to bring out the main ideas of the algorithm we develop. With abuse of notation, we continue to denote an instance of CSB with the same threshold as $(\bmu, \theta_s, Q)$, where $\theta_s \in (0, Q]$. Note that the threshold is the same, but the mean losses can be different across the arms.  Though $\theta_s$ can take any value in the interval $(0, Q]$, a threshold equivalent to $\theta_s$ can be confined to a finite set. The following lemma shows that a threshold equivalent lies in a set consisting of the $K$ elements. 
\begin{restatable}{lemma}{ThetaSet}
	\label{lem:thetaSet}
	Let $\theta_s \in (0,Q]$, $M=\min\{\lfloor Q/\theta_s \rfloor,$ $K\}$ and $\hat{\theta}_s=Q/M$. 
	Then $\theta_s$ and $\hat{\theta}_s$ are threshold equivalent. Further, $\hat{\theta}_s \in \Theta$ where $\Theta = \{ Q/K, Q/(K-1), \cdots, Q\}$. 
\end{restatable}

Let $M= \min\{\lfloor Q/\theta_s\rfloor, K\}$. When arms are sorted in the decreasing order of mean losses, we refer to the first $M$ arms as the \emph{top-}$M$ arms and the remaining arms as \emph{bottom-}$(K-M)$ arms. The optimal allocation with the same threshold $\theta_s$ is to allocate $\theta_s$ amount of resource to each of the \emph{top-}$M$ arms and allocate the remaining resources to the other arms. The detailed proof of \cref{lem:thetaSet} and all other missing proofs appear in \cref{sec:appendix}.

\cref{lem:thetaSet} shows that the candidates for the threshold equivalent $\hat{\theta}_s$ for any instance $(\bmu,\theta_s,Q)$ are finite. Once the threshold equivalent is known, the problem reduces to identifying the \emph{top-}$M$ arms and assigning resource $\hat{\theta}_s$ to each one of them to minimize the total mean loss. The latter part is equivalent to solving a Multiple-Play Multi-Armed Bandits problem, as discussed next.

After knowing the allocation equivalent, a learner's optimal policy is to allocate $\hat\theta_s$ fraction of resource among $M$ arms having the highest mean loss. As initially, mean losses are not known, empirical estimates of the losses can be used. When resource $\hat{\theta}_s$ is allocated to $M$ arm having the highest empirical losses, no loss is observed from them, but a loss of each of the remaining $K-M$ arms is observed (semi-bandits). In bandits literature, such problems where one can sample rewards (losses) from a subset of arms is known as the Stochastic Multiple-Play Multi-Armed Bandits (MP-MAB) problem. Thus once the learner identifies a threshold equivalent of $\theta$, the CSB problem is equivalent to solving an MP-MAB problem. We adapt the MP-TS algorithm \citep{ICML15_komiyama2015optimal} to our problem as it is shown to achieve optimal regret bound for Bernoulli distributions.

\subsubsection{Equivalence to Multiple-Play Multi-Armed Bandits}
The learner can play a subset of arms in each round known as superarm \citep{TAC1987_MultiPlayBandits_Anatharam} in the stochastic Multiple-Play Multi-Armed Bandits (MP-MAB) \cite{ICML15_komiyama2015optimal}. The size of each superarm is fixed (and known). The mean loss of a superarm is the sum of the means of its constituting arms. The learner plays a superarm in each round and then observes the loss from each arm played (semi-bandit feedback). The learner's goal is to play a superarm that has the smallest mean loss. A policy in MP-MAB selects a superarm in each round based on past information. The policy's performance is measured in terms of regret, defined as the difference between cumulative loss incurred by policy and that incurred by playing an optimal superarm in each round. Let $(\bmu, m) \in [0,1]^K \times \N_+$ denote an instance of MP-MAB where $\bmu$ denotes the mean loss vector, and $m \le K$ denotes the size of each superarm. Let $\PCSB_s \subset \PCSB$ denote the set of CSB instances with the same threshold for all arms. For any  $(\bmu,\theta_s, Q) \in \PCSB_s$ with $K$ arms and known threshold $\theta_s$, let $(\bmu, m)$ be an instance of MP-MAB with $K$ arms and each arm has the same Bernoulli distribution as the corresponding arm in the CSB instance with $m=K-M$, where $M=\min\{\floor{Q/\theta_s}, K\}$ as earlier. Let $\PMP$ denote the set of resulting MP-MAB problems and $f: \PCSB_s \rightarrow \PMP$ denote the above transformation.

Let $\pi$ be a policy on $\PMP$. We can use the policy $\pi$ for any $(\bmu,\theta_s, Q) \in \PCSB_s$ with known $\theta_s$ to select which set of arms to allocate resources. It is done as follows: In round $t$, let the information $(C_1, Y_1, C_2,Y_2, \ldots, C_{t-1}, Y_{t-1})$ collected from a CSB instance, where $C_r$ is the set of $K-M$ arms where no resource is allocated  in round $r$ and $Y_r$ is the samples observed from these arms. This information is given to policy $\pi$, which returns a set $C_t$ with $K-M$ elements in round $t$. Then all arms other than arms in $C_t$ are given resource $\theta_s$. Let this policy on $(\bmu,\theta_s, Q) \in \PCSB_s$ be denoted as $\pi^\prime$. Similarly, let $\beta^\prime$ be a policy on $\PCSB_s$ that can be adapted to yield a policy for $\PMP$ as follows: In round $t$, let the information $(M_1, Y_1, M_2,Y_2, \ldots, M_{t-1}, Y_{t-1})$ collected from an MP-MAB instance, where $M_r$ is the superarm played in round $r$ and $Y_r$ is the associated loss observed from each arms in $M_r$,  is given to the policy $\beta^\prime$ which returns a set $M_t$ of $K-M$ arms where no resources has to be applied. The superarm corresponding to $M_t$ is then played. Let this policy on $\PMP$ be denoted as $\beta$. Note that when $\theta_s$ is known, the mapping is invertible. Our next result gives regret equivalence between the MP-MAB problem and CSB problem with a known same threshold. 

\begin{restatable}{proposition}{RegretEquiST}
	\label{prop:RegretEquiST}
	Let  $f: \PCSB_s \rightarrow \PMP$ and $P=(\bmu,\theta_s, Q) \in \PCSB_s$ with known $\theta_s$. Then the regret of  policy $\pi^\prime$ on $P$ is same as the regret of policy $\pi$ on $f(P)$. Similarly, let $P^\prime=(\bmu,m) \in \PMP$, then the regret of a policy $\beta$ on $P^\prime$ is same as the regret of policy $\beta^\prime$ on $f^{-1}(P^\prime)$. Thus the set $\PCSB$ with a known $\theta_s$ is 'regret equivalent' to $\PMP$, i.e., $\Regret(\PCSB_s)=\Regret(\PMP)$. 
\end{restatable}

The above proposition suggests that any algorithm which works well for the MP-MAB problem also works well for the CSB problem once the threshold is known. Hence one can use MP-MAB algorithms like MP-TS \citep{ICML15_komiyama2015optimal} and ESCB \citep{NIPS15_combes2015combinatorial} after knowing the threshold equivalent of $\theta_s$. MP-TS uses Thompson Sampling, whereas ESCB uses UCB (Upper Confidence Bound) and KL-UCB type indices. One can use any one of these algorithms. But we adapt MP-TS to our setting as it gives better empirical performance and is shown to achieve optimal regret bound for Bernoulli distributed rewards (losses). We next discuss the lower bound for CSB instances with the same threshold.

\paragraph{Lower bound.} As a consequence of the above equivalence and one-to-one correspondence, a lower bound on MP-MAB is also a lower bound on the CSB instance with the same threshold. Therefore, the following lower bound given for any strongly consistent algorithm \cite[Theorem 3.1]{TAC1987_MultiPlayBandits_Anatharam} is also a lower bound on the CSB problem with the same threshold:
\begin{equation}
\label{eqn:LowerBound}
\lim_{T\rightarrow \infty} {\frac{\mathbb{E}[\Regret_T]}{\log T} \ge \sum_{i=1}^{M}  \frac{\mu_{i}-\mu_{M+1}}{d(\mu_{M+1}, \mu_{i})} },
\end{equation}
where $d(p,q)$ is the KL divergence between two Bernoulli distributions with parameter $p$ and $q$. Also note that we are in loss setting.

\subsubsection{Algorithm: \ref{alg:CSB-ST}}
We develop an algorithm named \ref{alg:CSB-ST} for solving the Censored Semi-Bandits problem having the same threshold for all arms. It exploits the result in \cref{lem:thetaSet}, to learn an allocation equivalent of threshold and regret equivalence established in \cref{prop:RegretEquiST} to minimize the regret using an MP-MAB algorithm. \ref{alg:CSB-ST} works as follows: It takes $\delta$ and $\epsilon$ as input, where $\delta$ is the confidence on the correctness of estimated allocation equivalent and $\epsilon$ is such that $\mu_K \ge \epsilon > 0$. The value of $\delta$ can be a function of horizon ($T$), e.g., $\delta=1/T$. We set the prior distribution for each arm's mean loss as the Beta distribution $\beta(1, 1)$. For each arm $i \in [K], ~S_i$ represents the number of rounds when the loss is $1$, and $F_i$ represents the number of rounds when the loss is $0$ whenever the arm $i$ receives resource above its threshold.

We initialize $\Theta=\{Q/K, Q/(K-1), \dots, Q\}$ as given in Lemma \ref{lem:thetaSet}. The elements of $\Theta$ are in increasing order, and each of them is a candidate for allocation equivalent of  $\theta_s$. We use the set $\Theta$ to find the threshold estimate $\hat{\theta}_s$, which is threshold equivalent to the underlying threshold $\theta_s$ with high probability (at least $1-\delta$) by doing a binary search over it. The search starts by taking $\hat{\theta}_s$ to be the middle element in $\Theta$. The variables $l,~u,$ and $j$ are maintained to keep track of the estimation of allocation equivalent. The variable $l$ represents the lowest index of the possible candidate for allocation equivalent, $u$ represents the largest index of the possible candidate for allocation equivalent, and $j$ represents the element of the set $\Theta$, which will be used as a threshold in the next round. Let $S_i(t)$ and $F_i(t)$ denote the values of $S_i$ and $F_i$ in the starting of the round $t$. In round $t$, a sample $\hat\mu_{t,i}$ is drawn from $\beta(S_i(t), F_i(t))$ for each arm $i \in [K]$, which is independent of other arms. The values of $\hat\mu_{t,i}$ are ranked in the decreasing order, and the \emph{top-}$(Q/\hat\theta_s)$ (denoted as set $A_t$) arms are allocated $\hat\theta_s$ amount of resource, and their losses are observed.

Before knowing allocation equivalent, if a loss is observed at any of the arms in the set $A_t$, it implies that $\hat{\theta}_s$ is an underestimate of allocation equivalent. Hence $\hat{\theta}_s$ and all the candidates smaller than the value of $\hat{\theta}_s$ in $\Theta$ are eliminated, and the binary search is repeated in the remaining half of the elements again by starting with the middle element. The loss and no-loss counts are also updated as $S_i = S_i + X_{t,i}, F_i = F_i + 1-X_{t,i}+Z_i$ for all arms. The variable $Z_i, ~\forall i\in [K]$ keeps track of how many times no loss is observed for arm $i$ before a loss is observed when the arm $i$ has allocated $\hat{\theta}_s$ amount of resources. The variable $Z_i, i \in [K]$ is maintained for each arm because the arms in $A_t$ may not be the same in each round. It allows us to distinguish the zeros observed when the arm receives over and under resource allocation. Once a loss is observed for any arm in set $A_t$, the variable $Z_i$ is reset to zero for all arms.

If no loss is observed for all arms in the set $A_t$, $Z_i$ is incremented by $1$ for each arm $i \in A_t$ and variable $C$ is incremented by $1$. The variable $C$ keeps track of the number of consecutive rounds for which no loss is observed on all the arms that are allocated $\hat{\theta}_s$ amount of resource. It changes to $0$ either after observing a loss or if no loss is observed for consecutive $W_\delta$ rounds, where the value of $W_\delta$ ensures $\hat{\theta}_s$ is an allocation equivalent with the probability of at least $1-\delta$. If $C$ equals $W_\delta$, then with high probability, $\hat{\theta}_i$ is possibly an overestimate of allocation equivalent. Accordingly, all the candidates larger than the current value of $\hat{\theta}_s$ in $\Theta$ are eliminated, and the binary search is repeated, starting with the middle element in the remaining half. Note that the current value of $\hat{\theta}_s$ is not eliminated because it is possible that $\hat{\theta}_s$ may be only upper bound for threshold. The value of $C$ as well as $Z_i,\; \forall i \in [K]$ are reset to $0$. Resetting $Z_i$ values to zero once the number of zeros observed reaches $W_\delta$ ensures that they do not add to $F_i$ values when the resources are over-allocated. After this, the loss and no-loss counts are updated as $S_i = S_i + X_{t,i}, F_i = F_i + 1-X_{t,i}$ for each arm $i \in [K] \setminus A_t$.

\begin{algorithm}[!ht] 
	\renewcommand{\thealgorithm}{\bf CSB-SK} 
	\floatname{algorithm}{}
	\caption{Algorithm for CSB problem having Same Threshold with Known Horizon and $\epsilon$}
	\label{alg:CSB-ST}
	\begin{algorithmic}[1]
		\State \textbf{Input:} $\delta, \epsilon$
		\State Set $W_\delta = {\log(\log_2(K)/\delta)}/({\log(1/(1-\epsilon))})$ and $\forall i \in [K]: S_i=1, F_i=1, Z_i=0$ 
		\State Initialize $\Theta$ as given in Lemma \ref{lem:thetaSet}, $C=0, l=1, u = K, j = \floor{(l+u)/2}$ 
		\For{$t=1,2,\ldots,$}
			\State Set $\hat{\theta}_s = \Theta[j]$ and $\forall i \in [K]: \hat{\mu}_{t,i} \leftarrow \beta(S_i, F_i)$
			\State $A_t \leftarrow$ set of \emph{top-}$({Q}/{\hat{\theta}_s})$ arms with the largest values of $\hat{\mu}_{t,i}$
			\State $\forall i \in A_t:$ allocate $\hat{\theta}_s$ resource and observe $X_{t,i}$
			\If{$j \ne u$} 
				\If{$X_{t,a} = 1$ for any $a \in A_t$} 
					\State Set $l=j+1, ~j = \floor{(l+u)/2}, C=0$
					\State $\forall i \in [K]$: set $S_i = S_i+X_{t,i}, F_i = F_i+1-X_{t,i} + Z_i, Z_i =0$
				\Else
					\State Set $C = C + 1$ and $\forall i \in A_t: Z_i = Z_i + 1$
					\State If $C = W_\delta$ then set $u=j, j = \floor{(l+u)/2}$, $C=0, \forall i \in [K]: Z_i =0$
					\State $\forall i \in [K]\setminus A_t: S_i = S_i + X_{t,i}, F_i = F_i + 1 - X_{t,i}$
				\EndIf
				\Else
					\State $\forall i \in [K]\setminus A_t: S_i = S_i + X_{t,i}, F_i = F_i + 1 - X_{t,i}$
			\EndIf
		\EndFor
	\end{algorithmic}
\end{algorithm}

Since $\Theta$ has $K$ elements, the search for an allocation equivalent of $\theta_s$ terminates in a finite number of rounds with high probability. Once this happens, the algorithm allocates resources among \emph{top-}$(Q/\hat\theta_s)$ arms (from Lemma \ref{lem:thetaSet}) in the subsequent rounds and observes losses from remaining arms, i.e., the losses are observed for $K-M$ arms (multiple-play) in each round, where $M = Q/\hat\theta_s$. Observe that the \emph{top-}$(Q/\hat\theta_s)$ arms correspond to top arms with the highest estimated means, which are generated from an associated beta distribution. Hence after finding the allocation equivalent of $\theta_s$, our algorithm is the same as MP-TS. We leverage this observation to adapt the regret bounds of MP-TS to our loss setting.

Once $\hat\theta_s$ is known, the mean losses vector $\bmu$ needs to be estimated. The resources can be allocated such that no losses are observed for maximum $M$ arms. As our goal is to minimize the mean loss, we have to select $M$ arms with the highest mean loss and then allocate $\hat\theta_s$ to each of them. It is equivalent to find $K-M$ arms with the least mean loss, then allocate no resources to these arms and observe their losses. These losses are then used for updating the empirical estimate of the mean loss of arms.

\subsubsection{Analysis of \ref{alg:CSB-ST}}
\label{sssec:sameThetaRegretBounds}
Note that when $\hat{\theta}_s$ is an underestimate, and no loss is observed for consecutive $W_\delta$ rounds, then $\hat{\theta}_s$ will be reduced, which leads to a wrong estimate of $\hat{\theta}_s$. To avoid this, we set the value of $W_\delta$ such that the probability of happening of such an event is upper bounded by $\delta$. The next lemma gives a bound on the number of rounds needed to find threshold equivalent for threshold $\theta_s$ with high probability.

\begin{restatable}{lemma}{sameThresholdEstRounds}
	\label{lem:sameThresholdEstRounds}
	Let $(\bmu, \theta_s, Q)$ be an CSB instance with same threshold, where $\mu_1 \geq  \epsilon>0$. Then with probability at least $1-\delta$, the number of rounds needed by \ref{alg:CSB-ST} to find the  threshold equivalent of  $\theta_s$ is upper bounded by 
	\begin{equation*}
	T_{\theta_{s}^k}\le \frac{\log(\log_2(K)/\delta)}{\log\left({1}/{(1-\epsilon)}\right)}\log_2(K).
	\end{equation*}
\end{restatable}

For instance $(\bmu,\theta, Q)$ and any feasible allocation $\ba \in \A$, we define $\nabla_{\ba} = \sum_{i=1}^K\mu_i\big(\one{a_i < \theta_i} - \one{a_i^\star < \theta_i}\big)$, $\nabla_{\max} = \max\limits_{\ba \in \A } \nabla_{\ba}$, and $\nabla_{\min} = \min\limits_{\ba \in \A } \nabla_{\ba}$. We are now ready to state the regret bound.

\begin{restatable}{theorem}{regretSameThreshold}
	\label{thm:regretSameThreshold}
	Let $\mu_K\geq \epsilon>0$, $W_\delta = {\log(\log_2(K)/\delta)}/{\log(1/(1-\epsilon))}$, $\mu_{M} > \mu_{M+1},$ and $T>T_{\theta_{s}^k}$. Set $\delta=T^{-(\log T)^{-\alpha}}$ in \ref{alg:CSB-ST} such that $\alpha >0$. Then the expected regret of \ref{alg:CSB-ST} is upper bounded by
	\begin{equation*}
		\EE{\Regret_T} \le W_\delta\log_2{(K)}\nabla_{\max}   + O\left((\log T)^{{2}/{3}}\right) + \sum_{i \in [M]} \frac{(\mu_i-\mu_{M+1} )\log {T}}{d( \mu_{M+1},\mu_i)}.
	\end{equation*}
\end{restatable}

The first term in the regret bound of Theorem \ref{thm:regretSameThreshold} corresponds to the regret due to the estimation of allocation equivalent, and the remaining regret corresponds to the expected regret incurred after knowing the allocation equivalent.
Observe that the assumption $\mu_K\ge\epsilon>0$ is only required to guarantee that the estimation of  allocation equivalent terminates in a finite number of rounds. This assumption is not needed to get the bound on expected regret after knowing allocation equivalent. The assumption $\mu_{M} > \mu_{M+1}$ ensures that Kullback-Leibler divergence in the regret bound is well defined. This assumption is also equivalent to assuming that the set of \emph{top-}$M$ arms is unique.

\begin{corollary}
	\label{cor:OptimalBoundST}
	The regret of \ref{alg:CSB-ST} is asymptotically optimal.
\end{corollary}

Setting $\delta=T^{-(\log T)^{-\alpha}}$ in \ref{alg:CSB-ST} for any  $\alpha>0$ leads to $W_\delta = O\left((\log T)^{1-\alpha}\right)$. Now the proof of Corollary \ref{cor:OptimalBoundST} follows by comparing the expected regret bound with the lower bound given in \cref{eqn:LowerBound}.

%% file: known_multiple.tex

We now consider a more general case, where the threshold may not be the same for all arms. We assume that the number of different thresholds are $n$. If $n=K$ then all thresholds are different. The first difficulty with this setup is finding an optimal allocation that needs not be just allocating resource to top $M$ arms. To see this, consider a  problem instance $(\bmu, \btheta, C)$ with $\bmu = (0.9,0.6,0.4)$, $\btheta = (0.6, 0.55, 0.45)$, and $Q=1$. The optimal allocation is $\ba^\star = (0, 0.55, 0.45)$ with no resource allocated to the top arm. Our next result gives the optimal allocation for an instance in $\PCSB$. Let $KP(\bmu,\btheta, Q)$ denote a $0$-$1$ knapsack problem with capacity $Q$ and $K$ items where item $i$ has weight $\theta_i$ and value $\mu_i$. 

\begin{restatable}{proposition}{diffThetaOptiSoln}
	\label{prop:diffThetaOptiSoln}
	Let $P=(\bmu,\btheta,Q) \in \PCSB$. Then the optimal allocation for $P$ is a solution of $KP(\bmu,\btheta, Q)$.
\end{restatable}

Observe that assigning $\theta_i$ resource to arm $i$ decreases the total mean loss by an amount $\mu_i$. As the goal is to allocate resources such that the total mean loss is minimized, i.e., $\min_{\ba \in \A}$  $\sum_{i\in[K]}\mu_i\one{a_i < \theta_i}$. It is equivalent to solving a 0-1 knapsack with capacity $Q$ where item $i$ has weight $\theta_i$ and value $\mu_i$. The second difficulty of having different thresholds is that the estimation of each arm's threshold is needed to be done separately. Unfortunately, we do not have a result equivalent of Lemma \ref{lem:thetaSet} so that the search space can be restricted to a finite set. We need to search over the entire $(0, Q]$ interval for each arm.

For an instance $P:=(\bmu,\btheta, Q)$, recall that $\ba^\star=(a_1^\star, \ldots, a_K^\star)$ denotes the optimal allocation. Let $r = Q - \sum_{i: a_i^\star \ge \theta_i}\theta_i$, where $r$ is the residual resources after the optimal allocation. Define $\gamma:=r/K$. Any instance with $\gamma = 0$ becomes a `hopeless' problem instance as the only vector that is the allocation equivalent of $\btheta$ is $\btheta$ itself, i.e., $a_i^\star=\theta_i, \;\forall i \in [K]$, which needs $\theta_i$ values to be estimated accurately to achieve optimal allocation. However, for $\gamma>0$, one can find the allocation equivalent with small errors in $\theta_i$ values; hence it can be estimated in a finite time as shown next result.
\begin{restatable}{lemma}{diffTheteEst}
	\label{lem:diffTheteEst}
	Let $\gamma = r/K$ and  $\forall i \in  [K]: \hat\theta_i \in [\theta_i, \ceil{\theta_i/\gamma} \gamma]$. Then $\hat{\btheta}$ is allocation equivalent of $\btheta$. 
\end{restatable}
The proof follows by an application of Theorem 3.2 in \cite{DO13_hifi2013sensitivity}, which gives conditions for two weight vectors $\btheta_1$ and $\btheta_2$ to have the same solution in $KP(\bmu,\btheta_1,Q)$ and $KP(\bmu,\btheta_2, Q)$ for fixed $\bmu$ and $Q$. 
The next definition describes when we can say that two thresholds are different.
\begin{definition}
	We say that two thresholds $\theta_i$ and $\theta_j$ are {\em different} if 
	$\ceil{{\theta_i}/{\gamma}} \ne \ceil{{\theta_j}/{\gamma}}$.
\end{definition}    

\cref{lem:diffTheteEst} and the above definition implies that two thresholds are different if they have different thresholds in the allocation equivalent vector $\hat{\btheta}$.

Once we estimate the allocation equivalent with accuracy such that the estimated $\hat{\btheta}$ is an allocation equivalent of $\btheta$, the problem is equivalent to solving the $KP(\bmu,\hat{\btheta}, Q)$ provided we learn $\bmu$. The learning $\bmu$ is equivalent to solving a Combinatorial Semi-Bandits \citep{ICML13_chen2013combinatorial,NIPS15_combes2015combinatorial,NeurIPS20_perrault2020statistical,ICML18_wang2018thompson} problem. Combinatorial Semi-Bandits is a generalization of MP-MAB, where one needs to identify a superarm (a subset of arms from a collection of subsets) such that the sum of reward/loss of the arms in the selected superarm is the highest/ lowest. The selected superarm's size in each round may not be the same in the Combinatorial Semi-Bandits problem. We could use an algorithm that works well for the Combinatorial Semi-Bandits, like SDCB \citep{NIPS16_chen2016combinatorial}, CTS \citep{ICML18_wang2018thompson}, and CTS-BETA \citep{NeurIPS20_perrault2020statistical} for solving the CSB problem with the known threshold vector. CTS and CTS-BETA use Thompson Sampling, whereas SDCB uses the UCB type index. 
Our following result gives regret equivalence between the Combinatorial Semi-Bandits and CSB problem with multiple thresholds.
\begin{restatable}{proposition}{MultiThetaEquivalence}
	\label{prop:MultiThetaEquivalence}
	The CSB problem with the known threshold vector $\btheta$ is regret equivalent to a Combinatorial Semi-Bandits where Oracle uses $KP(\bmu,\btheta, Q)$ to identify the optimal superarm.
\end{restatable}

\subsubsection{Algorithm: \ref{alg:CSB-MT} }
We develop an algorithm named \ref{alg:CSB-MT} for solving the Censored Semi-Bandits problem with multiple thresholds. It exploits the result of \cref{lem:diffTheteEst} and the regret equivalence established in \cref{prop:MultiThetaEquivalence} to learn a good estimate of the threshold for each arm and minimizes the regret using the existing algorithm for Combinatorial Semi-Bandits. 
\ref{alg:CSB-MT} works as follows: It takes $n, \delta, \epsilon$ and $\gamma$ as inputs, where $n$ be the number of different thresholds\footnote{If the number of thresholds is unknown then the value of $n$ is set to $K$ in \ref{alg:CSB-MT}. It is equivalent to assuming that all thresholds are different.}, $\delta$ is the confidence on the correctness of estimated allocation equivalent, $\epsilon$ is such that $\mu_K \ge \epsilon > 0$, and $\gamma$ is the K$^{th}$ fraction of the leftover resources after having an optimal allocation of resources. We initialize each arm's prior distribution as the Beta distribution $\beta(1, 1)$. For each arm $i \in [K], S_i$ represents the number of rounds when the loss is $1$, and $F_i$ represents the number of rounds when the loss is $0$ whenever the arm $i$ receives resource above its threshold. The variable $Z_i$ keeps the count of consecutive $0$ for the arm $i$ when allocated the required resource. $Z_i$ changes to $0$ either after observing a loss or if no loss is observed for consecutively $W_\delta$ rounds where the value of $W_\delta$ ensures $\hat{\btheta}$ is an allocation equivalent with the probability of at least $1-\delta$.

The algorithm needs to find a threshold vector that is allocation equivalent of $\btheta$ with high probability. It is achieved by ensuring that $\hat\theta_i \in [\theta_i, \ceil{\theta_i/\gamma}\gamma]$ for each $i \in [K]$ (\cref{lem:diffTheteEst}). The algorithm maintains the variables $\theta_{l,i}, \theta_{u,i}$, $\theta_{g,i}$, and $\hat{\theta}_{i}$ for the estimation of allocation equivalent, where $\hat{\theta}_{i}$ is the estimated value of ${\theta}_{i}$; $\theta_{u,i}$ and $\theta_{l,i}$ is the upper and lower bound of the search region for the threshold of arm $i$ respectively; and $\theta_{g,i}$ indicates whether the current estimate of the threshold lies in the interval $[\theta_i, \ceil{\theta_i/\gamma}\gamma]$ for arm $i$.  
The algorithm also keeps track of set $\Theta_n$ and variable $n_i$, where $\Theta_n$ is the set of estimated thresholds and $n_i$ is the index of arm whose threshold will be searched in the set $\Theta_n$. The set $\Theta_n$ is initialized as empty set whereas the value of $n_i$ is set to $1$ if $n<K$ otherwise $0$. The value of $n_i=0$ ensures that when all thresholds are different, then the threshold is estimated separately for each arm.

Let $S_i(t)$ and $F_i(t)$ denote the value of $S_i$ and $F_i$ at the start of round $t$. In round $t$, for each $i \in [K]$ an independent sample for estimated loss $(\hat\mu_{t,i})$ is drawn from $\beta(S_i(t), F_i(t))$. If there exists any arm whose threshold is not good, then the allocation equivalent needs to be estimated. We say that the threshold estimate of arm $i$ is good by checking the condition $\theta_{u,i}  - \theta_{l,i} \le \gamma$. If the condition satisfies, then the estimated threshold of the arm is within the desired tolerance, and it is indicated by setting $\hat\theta_{g, i}=1$; otherwise, it remains $0$. 

\begin{algorithm}[!ht]
	\small 
	\renewcommand{\thealgorithm}{\bf CSB-MK}  
	\floatname{algorithm}{}
	\caption{Algorithm for CSB problem having Multiple Threshold with Known Horizon and $\epsilon$}
	\label{alg:CSB-MT}
	\begin{algorithmic}[1]
		\State \textbf{Input:} $n, \delta, \epsilon, \gamma$
		\State Initialize: $\forall i \in [K]: S_i = 1, F_i = 1, Z_i =0, \theta_{l,i} = 0, \theta_{u,i} = Q, \theta_{g,i} = 0,  \hat\theta_{i} = Q/2$ 
		\State Set $\Theta_n = \emptyset,W_\delta = \log (K\log_2(\lceil 1 + Q/\gamma\rceil)/\delta)/\log(1/(1-\epsilon)),$ if $n<K$ then $n_i =1$ else $n_i=0$ 
		\For{$t=1,2, \ldots,$}
		\State $\forall i \in [K]: \hat{\mu}_{t,i} \leftarrow \text{Beta}(S_i, F_i)$
		\If{$\theta_{g,j} = 0$ for any $j \in [K]$}
		\If{$n < K$}
		\While{$\theta_{g,n_ i} = 1$}
			\State  Add $\hat\theta_{n_i}$ to $\Theta_n$ and set $n_i = n_i + 1$.  Sort $\Theta_n$ in increasing order
		\EndWhile
		\If{there exists no $j \in [|\Theta_n|]$ such that $\theta_{l,n_i} < \Theta_n[j] \le \theta_{u,n_i}$ or $\Theta_n = \emptyset$}
		\State  Set $\hat\theta_{n_i}=(\theta_{l,i}+\theta_{u,i})/2$ 
		\Else
		\State Set $l = \min\{k: \Theta_n[k] > \theta_{l,n_i}\}, u = \max\{k: \Theta_n[k] \le \theta_{u,n_i}\},$ and $j = \floor{(l+u)/2}$
		\State If $\Theta_n[j]=\theta_{u,n_i}$ then set $\hat\theta_{n_i} = \Theta_n[j]-\gamma$ else  $\hat\theta_{n_i} = \Theta_n[j]$
		\EndIf
		\EndIf
		
		\State $\forall i \in [K]\setminus \{n_i\}$: update $\hat\theta_{i}$ using Eq. \eqref{equ:updateTheta}. Allocate $\hat\theta_{i}$ resource to arm $i$ and observe $X_{t,i}$
		\For{$i = \{1,2,\ldots, K\}$}
		\If{$\theta_{g,i} = 0$ and $\hat\theta_{i}> \theta_{l,i}$}
		\State If $X_{t,i}=1$ then set $\theta_{l,i} = \hat\theta_{i}, S_i = S_i + 1, F_i = F_i + Z_i, Z_i =0$ else  $Z_i = Z_i + 1$
		\State If {$Z_i= W_\delta$} then set $\theta_{u,i} = \hat\theta_{i}, Z_i=0 $
		\State If $\theta_{u,i} - \theta_{l,i} \le \gamma$ then set $\theta_{g,i}=1$ and $\hat\theta_i = \theta_{u,i}$
		\ElsIf{$\hat\theta_{i} \le \theta_{l,i}$ or $\big\{ \theta_{g,i} = 1$ and $\hat\theta_{i}<\hat\theta_{u,i} \big\}$}
		\State Set $S_i = S_i+X_{t,i}$ and $F_i = F_i+1-X_{t,i}$ 
		\EndIf	
		\EndFor	
		
		\Else
		\State $A_t \leftarrow$ Oracle$\big( KP(\hat\bmu_{t}, \hat\btheta, C)\big)$ and $\forall i \in A_t:$ allocate $\hat\theta_{i}$ resource
		\State $\forall i \in [K]\setminus A_t:$ observe $X_{t,i}$, update $S_i = S_i+X_{t,i}$ and $F_i = F_i+1-X_{t,i}$
		\EndIf
		\EndFor
	\end{algorithmic}
\end{algorithm}

The threshold is estimated for each arm for finding a threshold equivalent vector. For this, the set $\Theta_n$ is updated by having all the estimated threshold from the arm having $\hat\theta_{g, i}=1$. The elements of the set $\Theta_n$ are sorted in increasing order, and the value of $n_i$ is incremented accordingly. By algorithm design, all the arms whose indices are smaller than the value of $n_i$ are having a good estimate of the threshold. For the arm whose index matches with the value of $n_i$, its threshold is first searched in the set $\Theta_n$ by doing a binary search over elements of the set $\Theta_n$.  If there is no element of the set $\Theta_n$ lies in between the values of lower and upper bound (element can be same as the value of upper bound) of the arm's threshold, then it implies that the threshold of the arm is not in the set $\Theta_n$. Hence, the threshold for arm is estimated using binary search in the interval $(\theta_{l,i}, \theta_{u,i}]$ by setting its value to $(\theta_{l,i}+\theta_{u,i})/2$ in the subsequent rounds.pose there exists an element of the set $\Theta_n$ in between the values of the lower and upper bound of the arm's threshold. In that case, the binary search is used to search the threshold in set $\Theta_n$ by finding the index of smallest $(l)$ and largest element $(u)$ in set $\Theta_n$ whose value is just larger than the lower bound and smaller than or equal to upper bound of the arm's threshold respectively. The element with index $\floor{(l+u)/2}$ is selected as threshold estimate. If the value of the selected threshold matches with the value of the upper bound of the arm's threshold, then it is decreased by $\gamma$ amount to ensure the estimate is indeed the good threshold value for the arm.

For all arms except the arm having index $n_i$, the resource allocation is updated after computing the following events:
\begin{align*}
B_i = \left\{ \frac{\theta_{l,i} + \theta_{u,i}}{2} \le Q - \sum_{\forall j < i:\theta_{g,j}=0} \left(\frac{\theta_{l,j} + \theta_{u,j}}{2} \right) \right\} \mbox{ and} \\
H_i = \left\{\theta_{u,i} \le Q - \sum_{\substack{j \in [K]:\theta_{g,j}=0 \\ \hat\theta_{j} \ne 0}} \hat\theta_{j} -  \sum_{\substack{k \in [K],\theta_{g,k}=1\\ \hat\mu_{t,k}/\theta_{u,k} > \hat\mu_{t,i}/\theta_{u,i} }}  \theta_{u,k} \right\}.
\end{align*}
The event $B_i$ is defined for all arm having a bad threshold estimate, i.e., $\theta_{g, i}=0$ and indicates whether the arm can get desired resources or not. The event $H_i$ is defined for all arms having good threshold estimates, i.e., $\theta_{g, i}=1$ and indicates if the arm can get the required resources or not. By construction, the event $B_i$ does not happen for arms having good threshold estimates, and the event $H_i$ does not happen for arms having a bad threshold estimate. The resources are first allocated among arms having bad threshold estimates to find the allocation equivalent as soon as possible. The leftover resource is allocated to arms with good threshold estimates to decrease the total loss. Among the arms having bad thresholds, the arm with the smallest index gets resources first, followed by the next smallest index. Whereas in the arms having good thresholds, the arms having the highest empirical loss to resource ratio, i.e., $\hat\mu_j/\hat\theta_{i}$ gets resource first, followed by second highest.  The $\hat\theta_{i}$ for arm $i$ is updated as follows:
\begin{align}
\label{equ:updateTheta}
\hat\theta_{i} = 
\begin{cases}
\hat\theta_{i}  							  &\mbox{if $H_i(t)$ happens,} \\
\frac{\theta_{l,i} + \theta_{u,i}}{2}  &\mbox{if $B_i$ happens,} \\
0 												   & \mbox{Otherwise.} 
\end{cases}
\end{align}

In round $t$, $\hat\theta_{i}$ amount of resources is allocated to arm $i \in [K]$ and then loss $X_{t,i}$ is observed. If a loss is observed from the arm $i$ that is having a bad threshold estimate ($\theta_{g, i}=0$) and $\hat\theta_i > \theta_{l, i}$, then it implies that $\hat\theta_i$ is an underestimate of $\theta_i$ and the lower end of search region (lower bound of threshold) is increased to $\hat\theta_i$, i.e., $\theta_{l,i}=\hat\theta_i$. 
The success and failure counts are also updated as $S_i = S_i + 1, F_i = F_i + Z_i$, and $Z_i$ is reset to $0$. If no loss is observed, then $Z_i$ is incremented by $1$. If no loss is observed after allocating $\hat\theta_i$ resources for successive $W_\delta$ rounds for arm $i$ with a bad threshold estimate, then it implies that $\hat\theta_i$ is overestimated. 
So, the upper bound of threshold is set to $\hat\theta_{i}$, i.e, $\theta_{u,i}=\hat{\theta}_{t,i}$ and $Z_i$ is reset to $0$.  After updating the lower or upper bound, the condition $\theta_{u, i}  - \theta_{l, i}\le \gamma$ is checked for knowing the goodness of the estimated threshold. If the condition holds, then the arm's threshold estimate is within desired tolerance, which is indicated by setting $\theta_{g, i}$ to 1 and $\hat\theta_i=\theta_{u, i}$ for the subsequent rounds. 
For arms either having resources less than lower bound of threshold ($\hat\theta_{i} \le \theta_{l, i}$) or having good threshold estimate with $\hat\theta_i < \theta_{u, i}$, their success and failure counts are updated as $S_i = S_i + X_{t,i}, F_i = F_i + 1 - X_{t,i}$.

Once we have good threshold estimates for all arms, we could adapt to any algorithm that works well for Combinatorial Semi-Bandits. We adapt the CTS-BETA \citep{NeurIPS20_perrault2020statistical} to our setting due to its better empirical performance. Oracle uses $KL(\hat\bmu_t, \hat\btheta, C)$ to identify the arms in the round $t$ where the learner has to allocate the required resource (denoted as set $A_t$). Each arm $i \in A_t$ has allocated $\hat\theta_i$ amount of resources. A loss $X_{t, i}$ is observed from each arm $i \in [K]\setminus A_t$ and then $S_i = S_i + X_{t,i}, F_i = F_i + 1 - X_{t,i}$ are updated.

\subsubsection{Analysis of \ref{alg:CSB-MT}}
\label{sssec:differentThetaRegretBounds}
The value of $W_\delta$ in \ref{alg:CSB-MT} is set such that the probability of estimated threshold does not lie in $[\theta_i, \ceil{\theta_i/\gamma}\gamma]$ for all arms is upper bounded by $\delta$.  The following lemma gives the upper bound on the number of rounds required to find the allocation equivalent for threshold vector $\btheta$ with a probability of at least $1-\delta$.

\begin{restatable}{lemma}{MultiTheta}
	\label{lem:MultiTheta}
	Let $n$ be the number of different thresholds, $A_{\theta_n}$ be the set of first $n$ arms having different thresholds, and  $(\bmu,\btheta, Q)$ be an instance of CSB such that $\gamma>0$ and $\mu_1 \geq  \epsilon>0$. Then with probability at least $1-\delta$, the number of rounds needed by threshold estimation phase of \ref{alg:CSB-MT} to find the allocation equivalent for threshold vector $\btheta$ is upper bounded by 
	\begin{equation*}
	T_{\theta_n} \le  \frac{\log( K \log_2(\ceil{1 +{Q}/{\gamma}})/\delta)} {\log(1/(1-\epsilon))} \left[\sum_{i \in A_{\Theta_n}} {\log_2 (\ceil{1 +  {Q}/{\gamma}})} + K{\log_2 (n +  1)}\right].
	\end{equation*}
\end{restatable}

Let $\nabla_{\max}$ and $\nabla_{\min}$ be defined as in Section \ref{sssec:sameThetaRegretBounds}. We redefine $W_\delta = \log (K\log_2(\lceil 1 + Q/\gamma\rceil)/\delta)/\log(1/(1-\epsilon))$. Let $\nabla_{i, \min}$ be the minimum regret for superarms containing arm $i$ and $K^\prime$ be the maximum number of arms in any feasible resource allocation. We are now ready to state the regret bound of \ref{alg:CSB-MT}.
\begin{restatable}{theorem}{regretDiffThreshold}
	\label{thm:regretDiffThreshold}
	Let $(\bmu,\btheta, Q)\in \PCSB$ such that $\gamma>0$, $\mu_K\geq \epsilon$, and $T>T_{\theta_n}$. Set $\delta=$ $T^{-(\log T)^{-\alpha}}$ in \ref{alg:CSB-MT} such that $\alpha >0$. Then the expected regret of \ref{alg:CSB-MT} is upper bounded as 
	\begin{align*}
		&\EE{\Regret_T} \le W_\delta\left[\sum_{i \in A_{\Theta_n}} {\log_2 (\ceil{1 +  {Q}/{\gamma}})} + K{\log_2 (n +  1)}\right] \nabla_{\max} + O\left(\sum_{i \in [K]}\frac{\log^2(K^\prime)\log T}{\nabla_{i, \min}} \right).
	\end{align*}
\end{restatable}

The first term of expected regret is due to the estimation of allocation equivalent. As it takes $T_{\theta_n}$ rounds to complete, the maximum regret due to the estimation of allocation equivalent is bounded by $T_{\theta_n}\nabla_{\max}$, where $\nabla_{\max}$ is the maximum regret that can be incurred in any round. The remaining terms correspond to the regret after knowing the allocation equivalent. The expected regret of \ref{alg:CSB-MT} is $O(K\log^2(K^\prime)\log T/\nabla_{\min})$, where $\nabla_{\min}$ is the minimum gap between the mean loss of optimal allocation and any non-optimal allocation. Since the regret scales as $\Omega(K\log T/ \nabla_{\min})$ for the combinatorial semi-bandits \citep{NeurIPS20_perrault2020statistical}, the regret of \ref{alg:CSB-MT} matches to the lower bound up to a logarithmic term.

%% file: unknown.tex

In this section, we propose algorithms for the CSB problem that do not need to know the time horizon and minimum mean loss. As in the previous section, we deal with cases of the same and different thresholds separately.

\subsection{Arms with Same Threshold}
\input{unknown_same}

\subsection{Arms with Different Threshold}
\input{unknown_different}

%% file: unknown_same.tex

First, we develop a Thompson-sampling based algorithm named \ref{alg:CSB-JS} for the CSB problem where all arms have the same threshold $\theta_s$. \ref{alg:CSB-JS} starts with equally distributing the resources among all the $K$ arms and continues to do the same in the following rounds until no loss is observed on any of the arms. Once the loss is observed from any of the arms, then it equally distributes the resources among top $K-1$ arms having the largest estimates of mean losses. The process is repeated till no loss is observed from arms that have been allocated resources. Along the way, the algorithms identify the allocation equivalent of $\theta_s$ and also learns the optimal allocation of resources. 

The pseudo-code of the algorithm is given in \ref{alg:CSB-JS}. It works as follows: For each $i\in [K]$, the variables $S_i$ and $F_i$ are used to keep track of the number of rounds in which the loss is observed or not observed, respectively. No loss is only observed from arm $i$ when it receives at least $\theta_s$ amount of resource. The prior loss distribution of each arm is set as the Beta distribution $\beta(1, 1)$ by initializing $S_i=1$ and $F_i=1$. For each arm $i \in [K]$, let $S_i(t)$ and $F_i(t)$ denote the values of $S_i$ and $F_i$ at the starting of round $t$. In every round $t$, a sample $\hat\mu_i$ is drawn for each arm $i \in [K]$ from $\beta(S_i(t), F_i(t))$ independent of everything else.  Then the top-$L$ arms having the largest empirical mean loss (denoted as set $A_t$) is selected to distribute the resources equally. The value of $L$ is initialized by $K$.

\begin{algorithm}[!ht] 
	\renewcommand{\thealgorithm}{\bf CSB-SU}
	\floatname{algorithm}{}
	\caption{Algorithm for CSB with Same threshold with Unknown parameters}
	\label{alg:CSB-JS}
	\begin{algorithmic}[1]
		\State Set $L=K, S_i = 1, F_i = 1, Z_i=0 ~~\forall i \in [K]$ 
		\For{$t=1, 2, \ldots$}
		\State $\forall i \in [K]: \hat{\mu}_i(t) \leftarrow \beta(S_i, F_i)$
		\State $A_t \leftarrow$ set of $L$ arms with the largest values of $\hat{\mu}_{t,i}$
		\State $\forall i \in A_t$: allocate ${Q}/{L}$ resources and observe $X_{t,i}$
		\If{$X_{t,j}=1$ for any $j \in A_t$}
		\State  Set $L = L-1$. $\forall i \in A_t:$ update $S_i = S_i+$ $X_{t,i}, F_i = F_i+1-X_{t,i} + Z_i$. $\forall j \in [K]:$ set $Z_j=0$
		\Else
		\State $\forall i \in A_t:$ update $Z_i= Z_i+1$
		\EndIf
		\State $\forall i \in [K]\setminus A_t:$  update $S_i = S_i+X_{t,i}$, $F_i = F_i+1-X_{t,i}$
		\EndFor
	\end{algorithmic}
\end{algorithm}

If a loss is observed on any arms in the set $A_t$, then it implies that the current value of $\hat{\theta}_s$ is an underestimate of $\theta_s$. Hence $L$ is decreased by $1$ and then the success and failure counts are also updated as $S_i = S_i + X_{t,i}, F_i = F_i + 1-X_{t,i}+Z_i$ for each arm $i \in A_t$, and for the all $j \in [K]$, $Z_j$ is reset to $0$. The variables $(Z_i:\forall i\in [K])$ keep track of how many times no loss is observed for arms in the set $A_t$ before a loss is observed for any of arm in set $A_t$. Its value is reset to zero for all arms once a loss is observed for any arm in $A_t$. The variable $Z_i, i\in [k]$ is useful to distinguish between the loss due to randomness when resources are under-allocated and no loss due to over-allocation of resources. If no loss is observed for all arms in the set $A_t$, then $Z_i$ is incremented by $1$ for each arm $i \in A_t$. The values of $S_i$ and $F_i$ are updated for each arm where no resources are allocated.

Since there are only $K$ possible candidates for allocation equivalent, the allocation equivalent for $\theta_s$ is found in the finite number of rounds. Once allocation equivalent is known, the algorithm allocates resources equally among top-$M$ arms (\cref{lem:thetaSet}) in the subsequent rounds and observes loss samples for the remaining $K-M$ arms. The selected arms correspond to top-$M$ arms with the highest estimated mean losses. Hence after an allocation equivalent of $\theta_s, $ is reached, in each round, samples from the $K-M$ arms are observed, which corresponds to selecting the $K-M$ arms with the smallest means. \ref{alg:CSB-JS} is the same as MP-TS that plays $K-M$ arms in each round and aims to minimize the sum of mean losses incurred from $K-M$ arms. We exploit this observation to adapt the regret bounds of MP-TS.

\subsubsection{Analysis of \ref{alg:CSB-JS}}
\label{ssec:analysisCSB_JS}
Let $T_{\theta_s}$ denote number of rounds required to find an allocation equivalent of $\theta_s$. The first result gives the upper bounds on expected value of $T_{\theta_s}$.
\begin{restatable}{lemma}{sameThetaEstRounds}
	\label{lem:sameThetaEstRounds}
	Let $M$ be the number of arms in the optimal allocation. For CSB problem instance $(\bmu,\theta_s, Q)$, the expected number of rounds needed by \ref{alg:CSB-JS} to find an allocation equivalent for threshold $\theta_s$ is upper bounded as
	\begin{equation*}
	\EE{T_{\theta_s}} \le \sum_{L=M + 1}^{K} \frac{1}{1 - \Pi_{i \in [K]/[K-L]}(1-\mu_i)}.
	\end{equation*}	
\end{restatable}

\noindent
Let $\nabla_{\max}$ be defined as in Section \ref{sssec:sameThetaRegretBounds}. We are now ready the state the regret bounds.
\begin{restatable}{theorem}{regretJointSameThreshold}
	\label{thm:regretJointSameThreshold}
	Let $(\bmu,\theta_s, Q)$ be the CSB problem instance with same threshold, $\mu_{M} > \mu_{M+1}$ and $T>T_{\theta_s}$. Then the expected regret of \textnormal{\ref{alg:CSB-JS}} is upper bound as
	\begin{align*}
	\EE{\Regret_T} &\le \sum_{L=M + 1}^{K} \frac{\Delta_{\max}}{1 - \Pi_{i \in [K]/[K-L]}(1-\mu_i)} + O\left(\sum_{i =1}^{M} \frac{(\mu_i - \mu_{M+1})\log {T}}{d(\mu_{M+1},\mu_i)}\right).
	\end{align*}
\end{restatable}

 The proof of \cref{lem:sameThetaEstRounds} follows by deriving the number of rounds required to observe a sample of `$1$' from a set of independent Bernoulli random variables. Whereas for \cref{thm:regretJointSameThreshold}, the first term in the regret bound corresponds to the expected regret incurred due to the estimation of allocation equivalent. The second term in the regret bound corresponds to the expected regret due to the MB-MAB based regret minimization algorithm MP-TS \citep{ICML15_komiyama2015optimal}. The assumption $\mu_{M} > \mu_{M+1}$ ensures that Kullback-Leibler divergence in the regret bound is well defined. 

\begin{corollary}
	\label{cor:OptimalBoundJointST}
	The regret of \textnormal{\ref{alg:CSB-JS}} is asymptotically optimal.
\end{corollary}
The proof follows by comparing the asymptotic regret bound of \ref{alg:CSB-JS} with the lower bound of regret given in \cref{eqn:LowerBound}.

%% file: unknown_different.tex

In this section, we develop an algorithm named \ref{alg:CSB-JD} for the CSB problem where the thresholds may not be the same. It exploits \cref{lem:diffTheteEst} to find allocation equivalent. \ref{alg:CSB-JD} works as follows: It takes $\gamma$ as input. We initialize each arm's prior distribution as the Beta distribution $\beta(1, 1)$. For each arm $i \in  [K]$, algorithm maintains a variable $L_i$ and set $Z_i$. The variable $L_i$ is the lower bound of the threshold for arm $i$, set $Z_i$ keeps count of the number of time no loss is observed from the arm $i$ for different resource allocations, and $Z_i[\hat\theta_i]$ represents the count of no losses for resource allocation $\hat\theta_i$ to arm $i$. The value of $L_i$ is initially set to $0$, and set $Z_i$ is initialized as an empty set. The set $Z_i[\cdot]$ plays a similar role as to the variable $Z_i$ in \ref{alg:CSB-JD}; however, it needs to store the counts for different resource allocations.

\begin{algorithm}[!ht] 
	\renewcommand{\thealgorithm}{\bf CSB-DU} 
	\floatname{algorithm}{}
	\caption{Algorithm for CSB with Different threshold with Unknown parameters}
	\label{alg:CSB-JD}
	\begin{algorithmic}[1]
		\State \textbf{Input:} $\gamma$
		\State $\forall i \in [K]:$ set $S_i = 1, F_i = 1, L_i=0$, and $Z_i = \phi$
		\For{$t=1, 2, \ldots$}	
		\State $\forall i \in [K]: \hat\mu_{i} \leftarrow \beta(S_i, F_i)$.
		\If{$Q - \sum_{i \in [K]} (L_i + \gamma) \ge 0$}
		\State Compute $\hat\btheta$ using \eqref{equ:updateAllocation} and set $A_t \leftarrow [K]$
		\Else
		\State $A_{t} \leftarrow KP( \hat{\bmu}, \hat\btheta, Q)$ where $\hat\theta_{i}=L_i + \gamma$
		\EndIf
		
		\For{$i \in A_t$}
		\State Assign $\hat\theta_{i}$ resources to arm $i$ and observe $X_{t,i}$
		\If {$X_{t,i} = 1$} 
		\State If $L_i < \hat\theta_{i}$ then change $L_i = \hat\theta_{i}$ 
		\State Update $S_i \leftarrow S_i + 1$, $F_i \leftarrow F_i + \sum_{\hat\theta_i \le L_i} Z_i[\hat\theta_i]$, and $\forall \hat\theta_i \le L_i:$ set $Z_i[\hat\theta_i] = 0$
		\Else
		\State If {$\hat\theta_{i}$ is not in $Z_i$} then add $Z_i[\hat\theta_{i}] = 1$ to $Z_i$ otherwise update $Z_i[\hat\theta_{i}] = Z_i[\hat\theta_{i}] + 1$
		\EndIf
		\EndFor
		\State $\forall i \in [K]\setminus A_t:$ observe $X_{t,i}$ and update $S_i = S_i + X_{t,i}, F_i = F_i + 1 - X_{t,i}$
		\EndFor
	\end{algorithmic}
\end{algorithm}

Let $S_i(t)$ and $F_i(t)$ denote the value of $S_i$ and $F_i$ at beginning of the round $t$. In round $t$, for each $i \in [K]$, an independent sample $(\hat\mu_{t,i})$ is drawn from $\beta(S_i(t), F_i(t))$. Initially, the value of the lower bound of the threshold for each arm is set to $0$. At the start, the resources are equally distributed among the arms. In the subsequent rounds, it is incremented by an amount of $\gamma$ for arms on which a loss is observed while uniformly distributing leftover resources among other arms as follows:
\begin{align}
\label{equ:updateAllocation}
\theta_{i} = \begin{cases}
L_i + \gamma & \text{if } L_i \ne 0 \\
Q_l/L_0       & \text{otherwise}
\end{cases}
\end{align}
where $Q_l := Q - \sum_{i \in [K]: L_i \ne 0} (L_i + \gamma)$ are the leftover resources and $L_0 := |\{i: L_i =0\}|$ is the number of arms whose lower bound of threshold is still $0$. Allocating resources equally among arms leads to a better initial lower bound on thresholds.  This process is continued until all the arms can get the required resources.

If resources are not enough for all arms, then the set of arms is selected by solving $KP(\hat{\bmu}, \hat{\btheta}, Q)$ problem (denoted as set $A_t$). Each arm $i \in A_t$ has given resource $\theta_{i}=L_i + \gamma$ and a sample $X_{t,i}$ is observed. If a loss is observed, then it implies that the arm is under-allocated. Accordingly, the lower bound of the threshold for that arm is updated. The success and failure counts are also updated as $S_i = S_i + 1, F_i = F_i + \sum_{\hat\theta_i \le L_i} Z_i[\hat\theta_i]$, and values of set $Z_i[\hat\theta_i]$ with $\hat\theta_i \le L_i$ are changed to $0$. If no loss is observed for arms having required resources and $Z_i[\hat\theta_i]$ is not in set $Z_i$, then add $Z_i[\hat\theta_i]$ to set $Z_i$ with value $1$; otherwise, increment $Z_i[\hat\theta_i]$ by $1$. The success and failure counts are also updated for each arm $i \in [K]\setminus A_t$ as $S_i = S_i + X_{t,i}$ and $F_i = F_i + 1 - X_{t,i}$.

\subsubsection{Analysis of \ref{alg:CSB-JD}} 
Let $T_{\theta_d}$ denote the number of rounds required to find an allocation equivalent of $\btheta$. Our following result gives an upper bound on the expected value of $T_{\theta_d}$.
\begin{restatable}{lemma}{diffThetaEstRounds}
	\label{lem:diffThetaEstRounds}
	For CSB problem instance $(\bmu,\btheta, Q)$ with $\gamma > 0$, the expected number of rounds needed by \ref{alg:CSB-JD} to find an allocation equivalent vector for $\btheta$ is upper bounded as
	\begin{equation*}
	\EE{T_{\theta_d}} \le \sum_{i \in [K]: \mu_i \ne 0} \floor{\frac{\theta_i}{\gamma}} \left( \frac{1}{\mu_i} \right).
	\end{equation*}
\end{restatable}
The lower bound of the threshold for an arm having zero mean loss remains $0$. Therefore, when resources are not enough, \ref{alg:CSB-JD} allocate only $\gamma$ amount of resources to such arms. Let $\nabla_{\max}$, $\nabla_{i,\min}$, and $K^\prime$ be the same as in Section \ref{sssec:differentThetaRegretBounds}. We are now ready to state the regret bound. 
\begin{restatable}{theorem}{regretJointDiffThreshold}
	\label{thm:regretJointDiffThreshold}
	Let $(\bmu,\btheta, Q)$ be the CSB problem instance with $\gamma > 0$  and $T>T_{\theta_d}$. Then the expected regret of \textnormal{\ref{alg:CSB-JD}} is upper bound as
	\begin{align*}
		\EE{\Regret_T} \le &\sum_{i \in [K]: \mu_i \ne 0} \floor{\frac{\theta_i}{\gamma}} \left( \frac{1}{\mu_i} \right)\Delta_{\max} + O\left(\sum_{i \in [K]}\frac{\log^2(K^\prime)\log T}{\nabla_{i,\min}} \right).
	\end{align*}
\end{restatable}
The first term of expected regret is the regret incurred due to the estimation of allocation equivalent. The expected number of rounds needed to find the allocation equivalent is given by \cref{lem:diffThetaEstRounds}. The second term corresponds to the expected regret due to the combinatorial semi-bandits algorithm CTS-BETA \citep{NeurIPS20_perrault2020statistical}.

\paragraph{Anytime Algorithm for CSB problem with Multiple Thresholds.}
Anytime algorithms maintain the lower bound of thresholds and linearly increase resources after observing a loss for current resource allocation. These algorithms do not tell us whether the estimated threshold is good or bad. Hence it is not possible to maintain a set of good thresholds as done in \ref{alg:CSB-MT}. Suppose it is possible to maintain a list of possible candidates for thresholds. In that case, the resources are allocated accordingly among arms, which may not have a good threshold estimate. Since the anytime algorithms do not handle the over-estimation problem, there is no way to reduce the over-allocated resources to an arm as done by \ref{alg:CSB-MT} which waits for a certain number of rounds before reducing resources. We observe that anytime algorithms are not possible when considering the CSB problems in the reward setting. More discussion about this can be found in \cref{sec:num}, where we discuss the application of CSB setup for the Network Utility Maximization.

\paragraph{Anytime Algorithms versus Horizon-Dependent Algorithms.}
The significant difference between horizon dependent algorithms and anytime algorithms is the way they estimate the threshold vector. After knowing the allocation equivalent for the threshold vector, the algorithms work similarly. The horizon dependent algorithms use binary search and wait for a fixed number of rounds with over-allocated resources. In contrast, resources are increased linearly after observing a loss by anytime algorithms. Anytime algorithms perform poorly for CSB problems with different thresholds as the search space for allocation equivalent can be very large, but perform better for CSB problems with the same threshold, where the search space is small. \cref{table:regretComparision} summarizes the number of rounds taken for threshold estimation by proposed algorithms.

\begin{table}[!ht]
	\centering
	\setlength\tabcolsep{1pt}
	\setlength\extrarowheight{5pt}
	\begin{tabular}{ |c|c|c|} 
		\hline
		\diagbox[width=0.23\textwidth]{\bf Cases}{\bf $\boldsymbol{{\text{Rounds}}}$}
		&\thead{With Known Parameters $(\epsilon, \delta) $ \\ (Rounds with High Probability) }&\thead{With Unknown Parameters \\ (Expected Rounds)} \\ 
		\hline
		Same Threshold & $\frac{\log(\log_2(K)/\delta)}{\log\left({1}/{(1-\epsilon)}\right)} \log_2(K)$ & $\sum\limits_{L=M + 1}^{K} \frac{1}{1 - \Pi_{i \in [K]/[K-L]}(1-\mu_i)}$ \\ 
		\hline
		Different Threshold & $KW_\delta {\log_2\left(\ceil{1 +  {Q}/{\gamma}}\right)}$ & $\sum\limits_{i \in [K]:\mu_i \ne 0}\floor{\frac{\theta_i}{\gamma}} \left( \frac{1}{\mu_i} \right)$ \\ 
		\hline
		$1<n <K$ & $W_\delta \left(\sum\limits_{i \in A_{\Theta_n}} {\log_2 \ceil{1 +  \frac{Q}{\gamma}}} + K{\log_2 (n +  1)}\right)$ & -- \\ 
		\hline
	\end{tabular}
	\caption{Comparing upper bounds on the (expected) number of rounds needed to find allocation equivalent for the proposed algorithms. The value of $W_\delta$ used in the table is ${\log( K \log_2(\ceil{1 +{Q}/{\gamma}})/\delta)}/ {\log(1/(1-\epsilon))}$}
	\label{table:regretComparision}
\end{table}

%% file: experiment.tex

We empirically evaluate the performance of proposed algorithms on four synthetically generated instances. In instances I and II, the threshold is the same for all arms. In contrast, the thresholds vary across arms in Instance III and IV. The details are as follows:

\noindent
\textbf{Identical Threshold:} Both instance I and II have $K = 50, Q=15$ and $\theta_s=0.5$.  The mean loss of arm $i\in [K]$ is $x - (i-1)/100$. We set $x=0.5$ for instance I and $x=0.7$ for instance II. 

\noindent
\textbf{Different Thresholds:} Both Instance III and IV has $K = 10, Q=3$, and $\gamma=10^{-2}$. For Instance III, the mean loss vector is $\bmu = [0.9, 0.8, 0.42, 0.6, 0.5, 0.2, 0.1, 0.3, 0.7, 0.98]$ and the corresponding threshold vector is $\btheta=[0.65, 0.55, 0.3, 0.46, 0.37, 0.2, 0.07, 0.25, 0.3, 0.8]$. Whereas, Instance IV has the mean loss vector  $\bmu = [0.9, 0.8, 0.42, 0.6, 0.5, 0.2, 0.1, 0.3, 0.7, 0.98]$ and the corresponding threshold vector $\btheta=[0.55, 0.55, 0.3, 0.55, 0.55, 0.55, 0.3, 0.3, 0.3, 0.55]$.

The losses of the arm $i \in [K]$ are Bernoulli distributed with mean $\mu_i$. We repeated the experiment 100 times and plotted the regret with a 95\% confidence interval (the vertical line on each curve shows the confidence interval). 

\subsection{Performance of  Algorithms}
\begin{figure}[!ht]
	\centering
	\scriptsize
	\begin{subfigure}[b]{0.40\textwidth}	\includegraphics[width=\linewidth]{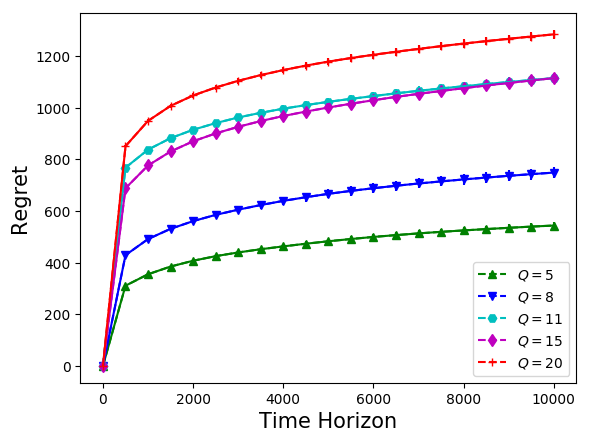}
		\caption{Varying resources in Instance II.}
		\label{fig:QSameTheta}
	\end{subfigure}\qquad
	\begin{subfigure}[b]{0.40\textwidth}
		\includegraphics[width=\linewidth]{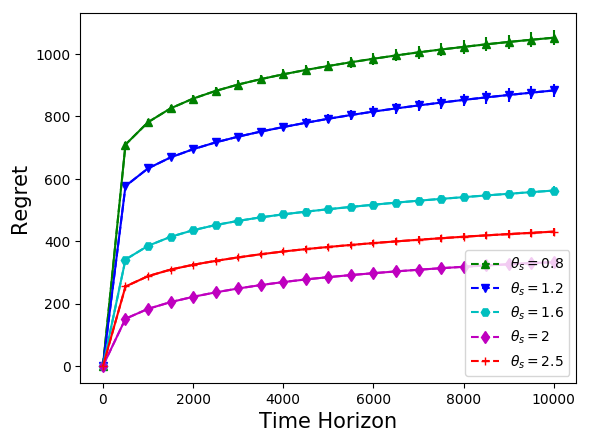}
		\caption{Varying value of same threshold.}
		\label{fig:TSameTheta}
	\end{subfigure}
	\caption{Regret of \ref{alg:CSB-ST} versus time horizon for the CSB problem with same threshold (Instance II).}
	\label{fig:knwon}
\end{figure}

In our first set of experiments, we empirically evaluate the performance of horizon dependent algorithms. First, we vary the amount of resource $Q$ for Instance II and observe the regret of \ref{alg:CSB-ST} as given in \cref{fig:QSameTheta}. We observe that when resources are small, the learner can allocate resources to a few arms but observes loss from more arms. On the other hand, when resources are more, the learner allocates resources to more arms but observes loss from fewer arms. Thus as resources increase, we move from semi-bandit feedback to bandit feedback. Therefore, regret increases with an increase in the amount of resources. Next, we only vary $\theta_s$ in Instance II, and the regret of \ref{alg:CSB-ST} for different value of same threshold $\theta_s$ is shown in \cref{fig:TSameTheta}. Similar trends are observed as the decrease in threshold leads to an increase in the number of arms that can be allocated resources and vice-versa. Therefore the amount of feedback decreases as the threshold decreases and leads to more regret. The empirical results also validate sub-linear regret bounds for the proposed algorithm. 

Since the regret depends on the optimal allocation and the amount of resources (threshold), the regret can vary with different resources (threshold) for the same optimal resource allocation. We can observe this behavior of regret in \cref{fig:QSameTheta} and \cref{fig:TSameTheta}. Note that horizon dependent algorithms need to know the lower bound on $\mu_i$ value and find the allocation equivalent with the probability of at least $1- \delta$. We set the lower bound on mean loss as $\epsilon=0.1$ and confidence parameter $\delta=1/T$ in the experiment that involves horizon dependent algorithms.

\begin{figure}[!ht]
	\centering
	\scriptsize
	\begin{subfigure}[b]{0.40\textwidth}	\includegraphics[width=\linewidth]{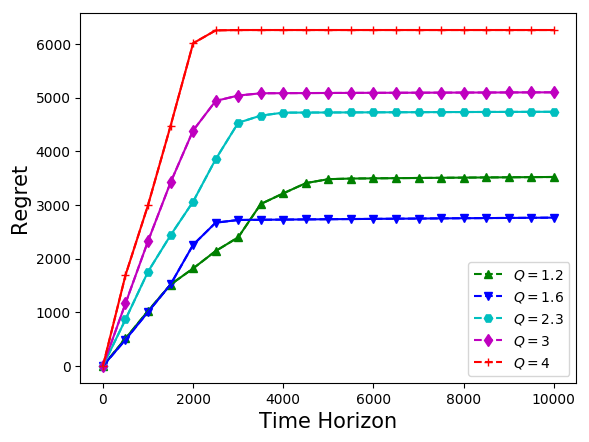}
		\caption{Varying resources in Instance III.}
		\label{fig:multiTheta1}
	\end{subfigure}\qquad
	\begin{subfigure}[b]{0.40\textwidth}
		\includegraphics[width=\linewidth]{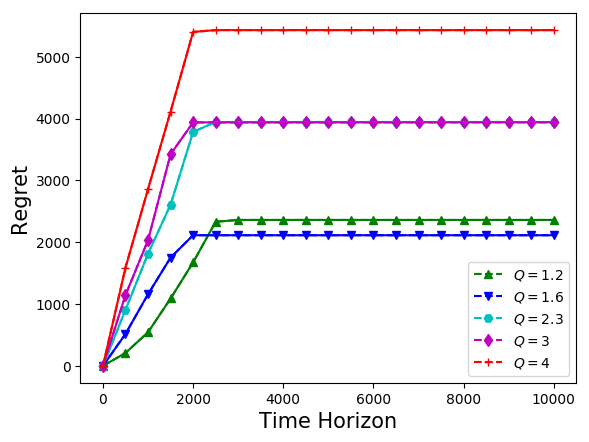}
		\caption{Varying resources in Instance IV.}
		\label{fig:multiTheta2}
	\end{subfigure}
	\caption{Regret of \ref{alg:CSB-MT} versus time horizon for the CSB problems with multiple thresholds.}
	\label{fig:multiTheta}
\end{figure}

In our next experiments, we change the available amount of resources in Instance III and IV. The regret of \ref{alg:CSB-MT} for the different amount of resources versus time horizon plots are shown in \cref{fig:multiTheta}. As expected, a similar behavior like \ref{alg:CSB-ST} is observed.

\begin{figure}[!ht]
	\centering
	\begin{subfigure}[b]{0.40\textwidth}
		\includegraphics[width=\linewidth]{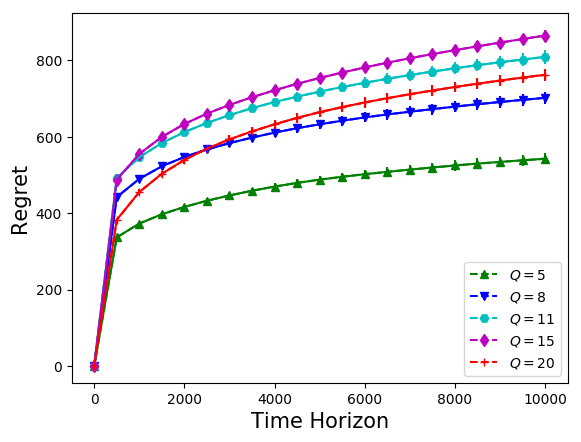}
		\caption{Varying resources in Instance II.}
		\label{fig:sameVaryQ}
	\end{subfigure}\qquad
	\begin{subfigure}[b]{0.40\textwidth}
		\includegraphics[width=\linewidth]{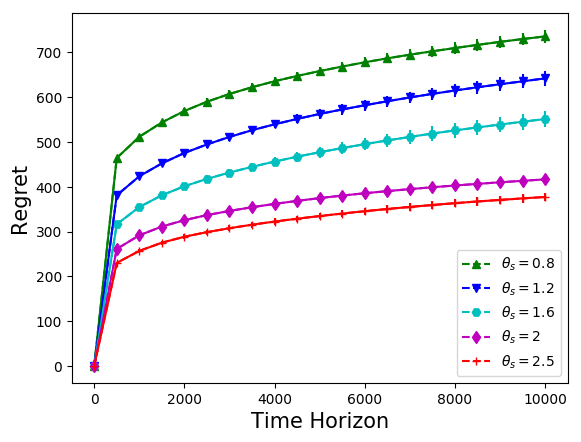}
		\caption{Varying value of same threshold.}
		\label{fig:sameVaryT}
	\end{subfigure}
	\caption{Regret of \ref{alg:CSB-JS} versus time horizon for the CSB problems with same thresholds.}
	\label{fig:SameTheta}
\end{figure}

\begin{figure}[!ht]
	\centering
	\scriptsize
	\begin{subfigure}[b]{0.40\textwidth}	\includegraphics[width=\linewidth]{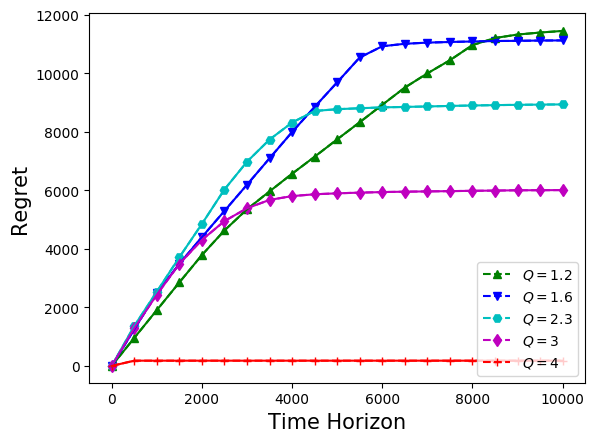}
		\caption{Varying resources in Instance III.}
		\label{fig:differemtTheta1}
	\end{subfigure}\qquad
	\begin{subfigure}[b]{0.40\textwidth}
		\includegraphics[width=\linewidth]{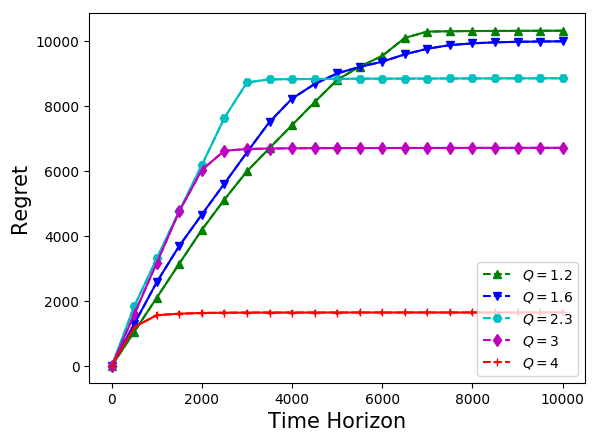}
		\caption{Varying resources in Instance IV.}
		\label{fig:differemtTheta2}
	\end{subfigure}
	\caption{Regret of \ref{alg:CSB-JD} versus time horizon for the CSB problems with different thresholds.}
	\label{fig:differemtTheta}
\end{figure}

We also run a similar set of experiments for anytime algorithms. The regret of anytime algorithms versus time horizon plots are shown in \cref{fig:SameTheta} and \cref{fig:differemtTheta}. As expected, we observe the same behavior as horizon dependent algorithms.

\subsection{Comparison between Algorithms}
\begin{figure}[!ht]
	\centering
	\begin{subfigure}[b]{0.40\textwidth}
		\includegraphics[width=\linewidth]{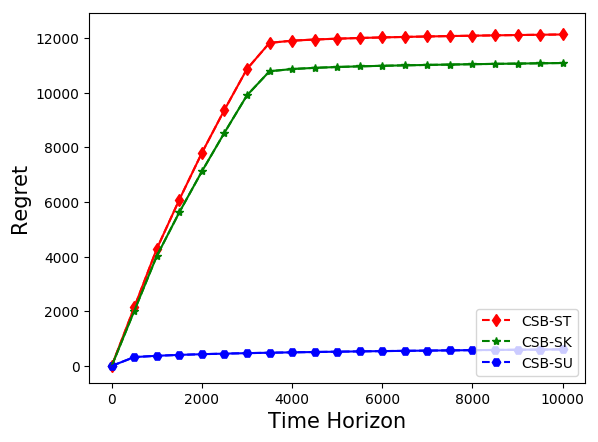}
		\caption{Comparison for Instance I}
		\label{fig:sameComp1}
	\end{subfigure}\qquad
	\begin{subfigure}[b]{0.40\textwidth}
		\includegraphics[width=\linewidth]{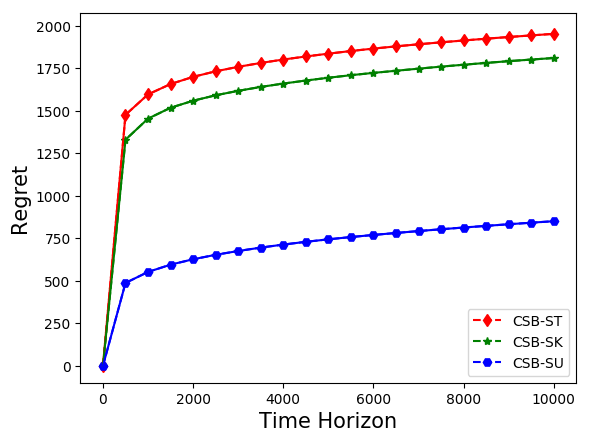}
		\caption{Comparison for Instance II}
		\label{fig:sameComp2}	
	\end{subfigure}
	\captionsetup{justification=centering}
	\caption{Comparing regret of \ref{alg:CSB-JS}, \ref{alg:CSB-ST}, and CSB-ST \citep{NeurIPS19_verma2019censored}.}
	\label{fig:CompSameTheta}
\end{figure}

We compare \ref{alg:CSB-JS}, \ref{alg:CSB-ST}, and state-of-the-art CSB-ST algorithm \citep{NeurIPS19_verma2019censored} for the CSB problems with the same threshold. Our algorithms outperforms CSB-ST for instance I and II as shown in \cref{fig:sameComp1} and \cref{fig:sameComp2}, respectively. Even though \ref{alg:CSB-ST} and CSB-ST use binary search for threshold estimation as compared to linear search in \ref{alg:CSB-JS}, there waiting delay with the overestimate of threshold leads to more rounds spend for the threshold estimation in considered CSB problems as compare to \ref{alg:CSB-JS}. Therefore, \ref{alg:CSB-JS} has the smallest regret than the other two algorithms.

\begin{figure}[!ht]
	\centering
	\begin{subfigure}[b]{0.40\textwidth}
		\includegraphics[width=\linewidth]{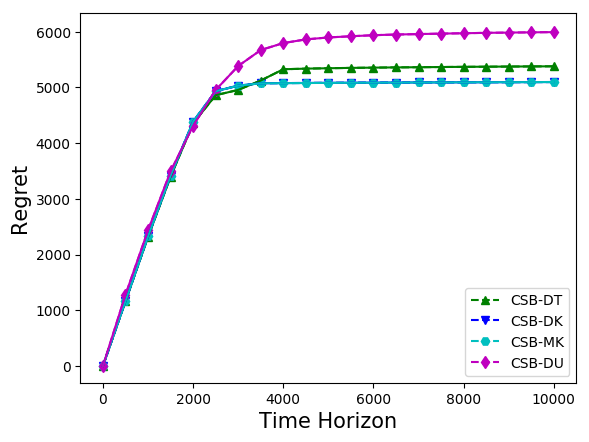}
		\caption{Comparison for Instance III}
		\label{fig:diffComp1}
	\end{subfigure}\qquad
	\begin{subfigure}[b]{0.40\textwidth}
		\includegraphics[width=\linewidth]{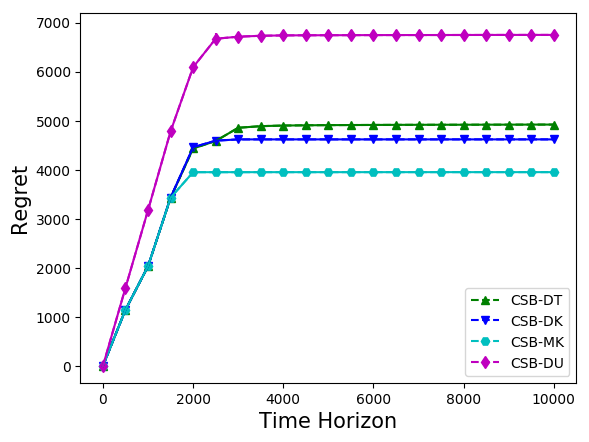}
		\caption{Comparison for Instance IV}
		\label{fig:diffComp2}	
	\end{subfigure}
	\captionsetup{justification=centering}
	\caption{\centering Comparing regret of \ref{alg:CSB-MT}, CSB-DK (assuming all thresholds are different which is equivalent to \ref{alg:CSB-MT} with $n=K$), \ref{alg:CSB-JD} and CSB-DT \citep{NeurIPS19_verma2019censored}.}
	\label{fig:diffTheta}
	
\end{figure}

We compare \ref{alg:CSB-MT}, CSB-DK (\ref{alg:CSB-MT} with $n=K$), \ref{alg:CSB-JD}, and state-of-the-art CSB-DT algorithm \citep{NeurIPS19_verma2019censored} for the CSB problems with different thresholds. \ref{alg:CSB-MT} and CSB-DT also uses binary search to estimate the threshold for each arm. These algorithms use the same threshold estimate for the fixed number of rounds, which depends upon the value of $\epsilon$ and $\delta$. The smaller the value of $\epsilon$, the more these algorithms wait for observing a loss and incur more regret as well. On the other hand, \ref{alg:CSB-JD} uses a linear search to estimate the threshold and does not need to know $\epsilon$ and $\delta$. As expected \ref{alg:CSB-MT} and CSB-DK outperform CSB-DT as shown in \cref{fig:diffComp2}. Whereas the performance of \ref{alg:CSB-MT} matches with CSB-DK in \cref{fig:diffComp1} as only two arms have the same threshold in Instance III. Since \ref{alg:CSB-JD} uses a linear search for threshold estimation, it needs more rounds to estimate allocation equivalent when the threshold has a larger search region than the algorithms that use binary search. Therefore, \ref{alg:CSB-JD} incurs more regret.

\subsection*{Computation complexity of 0-1 Knapsack with fractional weight and value}
Even though $KP(\bmu,\btheta, Q)$ is an NP-Hard problem; it can be solved by a pseudo-polynomial time algorithm\footnote{The running time of pseudo-polynomial time algorithm is a polynomial in the numeric value of the input whereas the running time of polynomial-time algorithms is polynomial of the length of the input.} using dynamic programming with the time complexity of O$(KQ)$. But such an algorithm for $KP(\bmu,\btheta, Q)$ works when the value and weight of items are integers. In the case of $\mu_i$ and $\theta_i$ are fractions, they need to be converted in integers with the desired accuracy by multiplying by large value $S$. The time complexity of solving $KP(S\bmu, S\btheta, SQ)$ is O$(KSQ)$ as a new capacity of Knapsack is $SQ$. Therefore, the time complexity of solving $KP(S\bmu, S\btheta, SQ)$ in each of the $T$ rounds is O$(TKSQ)$. 
Since solving the $0$-$1$ Knapsack problem is computationally expensive, we can solve it after $N$ rounds as the empirical mean losses do not change drastically in consecutive rounds in practice (except initial rounds). We have used $S = 10^4$ and $N = 20$ in our experiments involving the different thresholds.

%% file: conclusion.tex

We introduce a novel framework for resource allocation problems using a variant of semi-bandits and name it censored semi-bandits (CSB). In the CSB setup, the loss observed from an arm depends on the amount of resource allocated, and hence, it can be censored. We propose a threshold-based model where a loss from an arm is generated independently from a fixed and unknown distribution, but it is only observed when the allocated resource is below a threshold. The goal is to assign a given resource to arms such that total expected loss is minimized. We consider two variants of the problem, depending on whether or not the thresholds are the same across the arms. For the variant where thresholds are the same across the arms, we establish that its sub-problem is equivalent to the Multiple-Play Multi-Armed Bandit problem. For the second variant, where the threshold can depend on the arm, we establish that its sub-problem is equivalent to a more general Combinatorial Semi-Bandit problem. Exploiting these equivalences, we develop algorithms that enjoy optimal performance guarantees.  We also showcase the application of the CSB setup to stochastic network utilization maximization by extending the CSB setup to the reward maximization setting.

The CSB setting considered so far does not use any similarity metric between the arms. For example, in the case of the police patrol allocation, the nearby nodes may have similar parameters, and we may be able to make use of such spatial coherence. It will be interesting to integrate existing work on contextual bandits with the graph structure in the CSB setup. As we consider only one type of resource, it will be interesting to consider multi-type resources in the CSB setup. Another extension of the CSB setup is to relax the assumption that the lower bound on leftover resources after the optimal allocation is known. One can also extend the CSB setup from threshold type loss to continuous type loss functions and from the stochastic environment to the adversarial environment.

%% file: appendix.tex

\subsection*{Missing proofs from Section \ref{sec:known}}

\ThetaSet*
\proof{Proof.}
	The case $\floor{Q/\theta_s} \geq K$ is trivial. We consider the case $\floor{Q/\theta_s} < K$. By definition $M=\min\{\floor{Q/\theta_s},K\}$. We have $M \le Q/\theta_s$ and $\theta_s \le Q/M \doteq \hat{\theta}_s$. Hence $\hat{\theta}_s \ge \theta_s$. Therefore, $\hat{\theta}_s$ fraction of resource allocation for an arm has same reduction in the mean loss as $\theta_s$. Further, in both the instances $(\bmu, \theta_s, Q)$ and $(\bmu, \hat{\theta}_s, Q)$ the optimal allocations incur no loss from the {\emph top-}$M$ arms and the same amount of loss from the {\emph bottom-}$(K-M)$ arms. Hence the mean loss reduction for both the instances is same. This argument completes the proof of first part. As $M \in \{1, \ldots, K\}$ and $\hat{\theta}_s \le Q$, the possible value of $\hat{\theta}_s$ is only one of the elements in the set $\Theta = \{  Q/K, Q/(K-1), \cdots, Q\}$. 
\endproof

\sameThresholdEstRounds*
\proof{Proof.}
	When $\hat{\theta}_s < \theta_s$, it is possible that no loss is observed for $W_\delta$ consecutive rounds that leads to incorrect estimation of $\theta_s$. We want to set $W_\delta$ in such a way that the probability of occurring such event is upper bounded by $\delta$. This probability is bounded as follows:
	\begin{align*}
		&\Prob{\text{No loss is observed on $Q/\hat{\theta}_s$ arms for $W_\delta$ consecutive rounds at $\hat{\theta}_s<\theta_s$ (underestimate)}} \\
		&\qquad \le \prod_{i > K- Q/\hat{\theta}_s}^{K} (1 - \mu_i)^{W_\delta} \hspace{5mm}\text{\big(as $(1 - \mu_i)$ is the probability of not observing loss for arm $i$\big)}\\
		&\qquad \le \prod_{i > K-Q/\hat{\theta}_s}^{K} (1 - \epsilon)^{W_\delta} \hspace{8mm} \mbox{\big(as $\epsilon \ge \mu_i, ~\forall i \in [K]$\big)} \\
		&\qquad = (1 - \epsilon)^{\frac{QW_\delta}{\hat{\theta}_s}} \\	&\qquad  \le (1 - \epsilon)^{W_\delta}. \hspace{5mm}\text{\big(as $\hat{\theta}_s \le Q$\big)}
	\end{align*}
	Since we are using binary search and the set $\Theta$ has $K$ elements, the algorithm goes through at most $\log_2(K)$ underestimates of $\theta_s$. Let $I$ denote the set of indices of these underestimates in $\Theta$
	\begin{align*}
		&\Prob{\text{No loss is observed for consecutive $W_\delta$ rounds at any  underestimate of $\theta_s$}} \\
		&\qquad \le \sum_{i \in I}\Prob{\text{No loss is observed for consecutive $W_\delta$ rounds at the underestimate $\Theta(i)$}} \\
		&\qquad \le (1 - \epsilon)^{W_\delta}\log_2(K).
	\end{align*}
	As we are interesting in bounding the probability of making mistake by $\delta$, we get,
	\begin{align*}
	&(1 - \epsilon)^{W_\delta}\log_2(K) \le \delta \\
	\implies &(1 - \epsilon)^{W_\delta} \le \delta/\log_2(K).
	\end{align*}
	Taking log on both side of above equation, we get
	\begin{align*}
		&W_\delta\log(1 - \epsilon) \le \log(\delta/\log_2(K)) \\
		\implies &W_\delta\log\left(\frac{1}{1 - \epsilon}\right) \ge \log(\log_2(K)/\delta)\\ 
		\implies &W_\delta \ge \frac{\log(\log_2(K)/\delta)}{\log\left(\frac{1}{1 - \epsilon}\right)}.
	\end{align*}
	We set
	\begin{equation}
		W_\delta=\frac{\log(\log_2(K)/\delta)}{\log\left(\frac{1}{1 - \epsilon}\right) }.
	\end{equation}
	Hence, the minimum rounds needed to find a threshold that is an allocation equivalent with probability of at least $1-\delta$ is $W_\delta\log_2(K)$.
\endproof

\RegretEquiST*
\proof{Proof.}
	This result is an extension of  Proposition 1 in \cite{NeurIPS19_verma2019censored} to the case where $\theta_s \le Q$ instead of $\theta_s \le 1$.
	Let $\pi^\prime$ be a policy on $P:=(\bmu,\theta_s,Q) \in \PCSB_s$. The regret of policy $\pi^\prime$ on $P$ is given by
	\[\Regret_T(\pi^\prime, P)=\sum_{t=1}^T \left(\sum_{i=1}^K \mu_i\one{a_{t,i} < \theta_s} -\sum_{i=1}^K \mu_i\one{a^\star_i < \theta_s}\right), \]
	where $\ba^\star$ is the optimal allocation for $P$. Consider $f(P)=(\bmu,m) \in \PMP$ where $\bmu$ is same as in $P$ and $m=K-M$, where $M=\min\{\lfloor Q/\theta_s\rfloor,K\}$. The regret of policy $\pi$ on $f(P)$ is given by 
	\[\Regret_T(\pi, f(P))=\sum_{t=1}^T\left(\sum_{i \in M_t} \mu_i - \sum_{i>K-M}^{K}\mu_i\right),\]
	where $M_t$ is the superarm played in round $t$. Recall the ordering $\mu_1 \ge \mu_2 \ge \ldots \ge \mu_K$. It is clear that $\sum_{i=1}^K \mu_i\one{a^\star_i< \theta_s}=\sum_{i>M}^K\mu_i$. 
	Let $C_t$ be the set of arms where no resources are allocated by policy $\pi^\prime$ in round $t$. Since, loss only incurred from arms in the set $C_t$, we have $\sum_{i=1}^K \mu_i\one{a_{t,i} < \theta_s}=\sum_{i \in C_t}\mu_i$. By definition, the policy $\pi$ selects superarm $M_t=C_t$ in round $t$, i.e., set of arms returned by policy $\pi^\prime$ for which no resourced are applied. Hence $\sum_{i=1}^K \mu_i\one{a_{t,i}< \theta_s}=\sum_{i \in M_t} \mu_i $. This establishes the regret of policy $\pi^\prime$ on $P$ is same as regret of policy $\pi$ on $f(P)$ and hence, we get $\Regret(\PMP)\leq \Regret (\PCSB_s)$.
	Similarly, we can also establish the other direction of the proposition and get $\Regret(\PCSB_s)\leq \Regret (\PMP)$. Thus we conclude that $\Regret(\PCSB_s)= \Regret (\PMP)$.
\endproof

\noindent
We need the following results to prove the \cref{thm:regretSameThreshold}.
\begin{theorem}
	\label{thm:MPRegret}
	Let $\hat\theta_s$ be allocation equivalent of $\theta_s$ for instance $(\bmu,\theta_s,Q)$. Then, the expected  regret of \ref{alg:CSB-ST} for $T$ rounds after knowing the allocation equivalent is upper bounded as     
	\begin{align}
	\EE{\Regret_T} \le O\left((\log T)^{{2}/{3}}\right) + \sum_{i \in [M]} \frac{(\mu_i-\mu_{M+1} )\log {T}}{d( \mu_{M+1},\mu_i)}.
	\end{align}
\end{theorem}
\proof{Proof.}
	As $\hat\theta_s$ is the allocation equivalent of $\theta_s$, the instances $(\bmu,\theta_s,Q)$ and $(\bmu,\hat{\theta}_s,Q)$ have the same minimum loss. After knowing the allocation equivalent, the CSB problem with the same threshold is equivalent to solving an MP-MAB instance (\cref{prop:RegretEquiST}). Hence, we can directly apply Theorem 1 of \cite{ICML15_komiyama2015optimal} to obtain the regret bounds by setting $k=K-M$ and noting that we are in the loss setting and incur regret only when an arm $i \in [M]$ is in selected superarm.
\endproof

\begin{theorem}
	\label{thm:regretHighConf}
	With probability at least $1-\delta$, the expected cumulative regret of \ref{alg:CSB-ST} is upper bounded as
	\begin{equation*}
		\EE{\Regret_T} \le W_\delta\log_2{(K)} \nabla_{\max}  + O\left((\log T)^{{2}/{3}}\right) + \sum_{i \in [M]} \frac{(\mu_i-\mu_{M+1} )\log {T}}{d( \mu_{M+1},\mu_i)}.
	\end{equation*}
\end{theorem}

\proof{Proof.}
	The regret of \ref{alg:CSB-ST} can be divided into two parts: regret before knowing allocation equivalent and after knowing it. The threshold estimation completes in at most $W_\delta\log_2{(K)}$ rounds and returns a threshold which is an allocation equivalent with the probability of at least $1-\delta$. The maximum regret incurred for estimating allocation equivalent is $W_\delta\log_2{(K)} \nabla_{\max}$. The regret incurred after knowing allocation equivalent is given by \cref{thm:MPRegret}. Thus the expected regret of \ref{alg:CSB-ST} is the sum of regret incurred in the two parts and holds with the probability of at least $1-\delta$.  
\endproof

\noindent
We are now ready to give the proof of \cref{thm:regretSameThreshold}.
\regretSameThreshold*
\proof{Proof.}
	The regret bound follows from Theorem \ref{thm:regretHighConf} by setting $\delta=T^{-(\log T)^{-\alpha}}$ and unconditioning the expected regret incurred after knowing the allocation equivalent in \ref{alg:CSB-ST} .
\endproof

\diffThetaOptiSoln*
\proof{Proof.}
	Assigning $\theta_i$ fraction of resources to an arm $i$ reduces the total mean loss by amount $\mu_i$. Our goal is to allocate resources such that total mean loss is minimized, i.e., $\min\limits_{\ba \in \A}\sum_{i\in[K]}\mu_i\one{a_i < \theta_i}$. observe that the maximization version of same optimization problem is $\max\limits_{\ba \in \A}\sum_{i\in[K]}\mu_i\one{a_i \ge \theta_i}$ which is exactly same as solving a 0-1 knapsack with capacity $Q$ where item $i$ has value $\mu_i$ and weight $\theta_i$. 
\endproof

\diffTheteEst*
\proof{Proof.}
	Let $L^\star = \left\{i: a_i^\star < \theta_i\right\}$ and $r = Q - \sum_{i: a_i^\star \ge \theta_i}\theta_i$. If resource $r$ is allocated to any arm $i \in L^\star$, minimum value of mean loss will not change as $r < \min_{i \in L^\star} \theta_i$. If we can allocate $\gamma=r/K$ fraction of $r$ to each arm $i \in K$, the minimum mean loss still remains the same. If estimated threshold of every arm $i \in K$ lies in $[\theta_i, \ceil{\theta_i/\gamma}\gamma]$ then using Theorem 3.2 of \cite{DO13_hifi2013sensitivity}, $KP(\bmu,\btheta, Q)$ and $KP(\bmu, \hat\btheta, Q)$ has the same optimal solution because of having the same mean loss for both the problem instances.
\endproof

\MultiTheta*
\proof{Proof.}
	For any arm $i \in [K]$, we want $\hat{\theta}_i \in [\theta_i, \ceil{\theta_i/\gamma}\gamma]$. As $\theta_i \in (0,Q]$, we can divide interval $[0,Q]$ into a discrete set $\Theta \doteq \left\{0, \gamma, 2\gamma, \ldots, Q\right\}$ and note that $|\Theta| = \ceil{1+ {Q}/{\gamma}}$. As search space is reduced by half in each change of $\hat\theta_i$, the maximum change in $\hat\theta_i$ is upper bounded by $\log_2|\Theta|$ to make sure that $\hat{\theta}_i \in [\theta_i, \ceil{\theta_i/\gamma}\gamma]$. When $\hat{\theta}_i$ is underestimated and no loss is observed for consecutive $W_\delta$ rounds, a mistake happens by assuming that current allocation is an overestimate. We set $W_\delta$ such that the probability of estimating wrong $\hat{\theta}_i$ is small and bounded as follows:
	\begin{align*}
		&\Prob{\text{No loss is observed for consecutive $W_\delta$ rounds when $\hat{\theta}_i$ is underestimated}} \\
		&\qquad = (1 - \mu_i)^{W_\delta} \hspace{5mm}\text{\big(as $(1 - \mu_i)$ is the probability of not observing loss at arm $i$\big)}\\
		&\qquad \le (1 - \epsilon)^{W_\delta}. \hspace{6.8mm} \text{\big(since $\forall i \in [K]: \mu_i > \epsilon$\big)}
	\end{align*}
	Since we are doing binary search, the algorithm goes through at most $\log_2(|\Theta|)$ underestimates of $\theta_i$. Let $I$ denote the set of indices of these underestimates in $\Theta$
	\begin{align*}
		&\Prob{\text{No loss is observed for consecutive $W_\delta$ rounds when  $\hat{\theta}_i$ is underestimated}} \\
		&\qquad \le \sum_{i\in I}\Prob{\text{No loss is observed for consecutive $W_\delta$ rounds when  $\hat{\theta}_i$ is underestimated}} \\
		&\qquad \le (1 - \epsilon)^{W_\delta}\log_2(|\Theta|).
	\end{align*}
	Next, we will bound the probability of making mistake for any of the arm. That is given by
	\begin{align*}
	&\Prob{\exists i  \in [K], \hat\theta_i \in \Theta: \text{No loss is observed for consecutive $W_\delta$ rounds when  $\hat{\theta}_i$ is underestimated}} \\
	&~~ \le \sum_{i=1}^{K}\Prob{\exists \hat\theta_i \in \Theta: \text{No loss is observed for consecutive $W_\delta$ rounds when  $\hat{\theta}_i$ is underestimated}}  \\
	&~~ \le K(1 - \epsilon)^{W_\delta}\log_2(|\Theta|).
	\end{align*}
	
	As we are interested in bounding the above probability of making a mistake by $\delta$ for all arms, we have the following expression,
	\begin{align*}
		&K (1 - \epsilon)^{W_\delta}\log_2(|\Theta|) \le \delta\\
	 	\implies &(1 - \epsilon)^{W_\delta} \le \delta/K \log_2(|\Theta|).
	\end{align*}
	Taking log both side, we get
	\begin{align*}
		&W_\delta\log(1 - \epsilon) \le \log(\delta/K \log_2(|\Theta|))\\
	 	\implies & W_\delta\log\left({1}/{(1 - \epsilon)}\right) \ge \log(K \log_2(|\Theta|)/\delta)\\
		\implies &W_\delta \ge \frac{\log(K \log_2(|\Theta|)/\delta)}{\log\left({1}/{(1 - \epsilon)}\right) }.
	\end{align*}
	As $|\Theta| = \ceil{1+ Q/\gamma}$, we set 
	\begin{equation}
		W_\delta = \frac{\log(K \log_2 \ceil{1+ Q/\gamma}/\delta)}{\log\left({1}/{(1 - \epsilon)}\right) }.
	\end{equation}
	
	Therefore, the minimum number of rounds needed to find a threshold $\hat{\theta}_i$ for an arm $i$, which is an element of allocation equivalent vector with the probability of at least $1-\delta/K$ is upper bounded by $W_\delta\log_2 \ceil{1+ Q/\gamma}$. 
	
	Since $n$ is the number of different thresholds, there are $n$ different groups of arms where group $G_i$ consists of arms having the same estimated threshold $\hat{\theta}_i$ in the estimated allocation equivalent vector. We divided the number of rounds to know allocation equivalent into two parts. The first deals with the maximum number of expected rounds needed to find $n$ thresholds. In comparison, the second part deals with finding the good thresholds for remaining arms using known thresholds.
	
	Let consider the worst case where only one threshold is estimated at a time. Then the maximum expected rounds needed to estimate threshold associated with $G_i$ is $W_\delta \log_2\ceil{1 + Q/\gamma}$. Using this fact with definition of $A_{\theta_n}$, the maximum expected rounds needed to estimate $n$ thresholds is $W_\delta\sum_{i \in A_{\theta_n}} \log_2\ceil{1 + Q/\gamma}$. Once all $n$ thresholds are known then the threshold for any arm $k$ in remaining arms with non-zero mean loss need to search over $n$ possible values of thresholds and hence the expected number of rounds needed to its estimate is $W_\delta \log_2 (n+1)$. Therefore, the maximum number of rounds needed to estimate threshold for all arms in $A_{\theta_n}^c$ is $W_\delta\sum_{k \in A_{\theta_n}^c: \mu_k \ne 0} \log_2 (n+1)$ which further upper bounded by $W_\delta K\log_2 (n+1)$. With this argument, the proof is complete.
\endproof

\subsection*{Equivalence of CSB with different thresholds and Combinatorial Semi-Bandit }
In stochastic Combinatorial Semi-Bandits (CoSB), a learner can play a subset of $K$ arms in each round, also known as superarm, and observes the loss from each arm played \citep{ICML13_chen2013combinatorial,NIPS16_chen2016combinatorial,ICML18_wang2018thompson}. The size of a superarm can vary, and the mean loss of a superarm only depends on the mean of its constituent arms. The goal is to select a superarm that has the smallest loss. A policy in CoSB selects a superarm in each round based on past information. The performance of a policy is measured in terms of regret, defined as the difference between cumulative loss incurred by the policy and that incurred by playing an optimal superarm in each round. Let $(\bmu, \mathcal{I}) \in [0,1]^K \times 2^{[K]}$ denote an instance of CoSB, where $\bmu$ denote the mean loss vector and $\mathcal{I}$ denotes the set of superarms. Let $\PCSB_d \subset \PCSB$ denote the set of CSB instances with different thresholds. For any  $(\bmu,\btheta,Q) \in \PCSB_d$ with $K$ arms and known threshold $\btheta$, let $(\bmu, \mathcal{I})$ be an instance of CoSB with $K$ arms and each arm has the same Bernoulli distribution as the corresponding arm in the CSB instance. Let $\PCoSB$ denote set of resulting CoSB problems and $g: \PCSB_d \rightarrow \PCoSB$ denote the above transformation.

Let $\pi$ be a policy on $\PCoSB$. The policy $\pi$ can also be adapted for any $(\bmu,\btheta,Q) \in \PCSB_d$ with known $\btheta$ to decide which set of arms are allocated resource as follows: In round $t$, let information $(C_1, Y_1, C_2,Y_2, \ldots, C_{t-1}, Y_{t-1})$ collected from a CSB instance, where $C_s$ is the set of arms where no resource is applied and $Y_s$ is the samples observed from these arms, is given to $\pi$ which returns a set $C_t$. Then all arms other than arms in $C_t$ are given resource equal to their  estimated good threshold. Let this policy on $(\bmu,\btheta,Q) \in \PCSB_d$ is denoted as $\pi^\prime$. Similarly, a policy $\beta^\prime$ on $\PCSB_d$ can be adopted to yield a policy for $\PCoSB$ as follows: In round $t$, the information $(M_1, Y_1, M_2,Y_2, \ldots, M_{t-1}, M_{t-1})$, where $M_s$ is the superarm played in round $s$ and $Y_s$ is the associated loss observed from each arms in $M_s$, collected on an CoSB instance is given to the policy $\beta^\prime$. Then the policy $\beta^\prime$ returns a set $M_t$ where no resources has allocated. The superarm corresponding to $M_t$ is then played. Let this policy on $\PCoSB$ be denoted by $\beta$. Note that when $\btheta$ is known, the mapping is invertible. 
Our next result gives regret equivalence between the CoSB problem and the CSB problem with the known thresholds.

\MultiThetaEquivalence*
\vspace{-2.5mm}
\proof{Proof.}
	Let $\pi^\prime$ be a policy on $P:=(\bmu,\btheta,Q) \in \PCSB_d$. The regret of policy $\pi^\prime$ on $P$ is given by
	\[\Regret_T(\pi^\prime,P)=\sum_{t=1}^T \left(\sum_{i=1}^K \mu_i\one{a_{t,i}< \theta_i} -\sum_{i=1}^K \mu_i\one{a^\star_i < \theta_i}\right), \]
	where $\ba^\star$ is the optimal allocation for $P$. Consider $g(P)=(\bmu,\mathcal{I}) \in \PCoSB$ where $g: \PCSB_d \rightarrow \PCoSB$ and $\bmu$ is the same as in $P$ and $\mathcal{I}$ contains all superarms (set of arms) for which resource allocation is feasible. The regret of policy $\pi$ on $g(P)$ is given by 
	\vspace{-1.25mm}
	\[\Regret_T(\pi,g(P))=\sum_{t=1}^T\big(l(M_t,\bmu) - l(M^\star,\bmu)\big),\]
	where $M_t$ is the superarm played in round $t$,  $M^\star$ is optimal superarm, and $l$ returns mean loss for given superarm.
	The outcome of $l(M,\bmu)$ only depends on mean loss of constituents arms of the superarm $M$. In our setting, $l(M,\bmu) = \sum_{i \in M}\mu_i$ where $M= \left\{i: a_i <\theta_i \right\}$ for allocation $\ba \in \A$. It is clear that $\sum_{i=1}^K \mu_i\one{a^\star_i < \theta_i}=l(M^\star, \bmu)$. Let $C_t$ be the set of arms where no resource is allocated by $\pi^\prime$ in round $t$. Since, loss is only incurred for arms in the set $C_t$, we have $\sum_{i=1}^K \mu_i\one{a_{t,i}< \theta_s}=\sum_{i \in C_t}\mu_i$. By definition the policy $\pi$ selects superarm $M_t=C_t$ in round $t$, i.e., set of arms returned by $\pi^\prime$ for which no resourced are applied. Hence $\sum_{i=1}^K \mu_i\one{a_{t,i}< \theta_s}=\sum_{i \in M_t} \mu_i $. This establishes the regret of $\pi^\prime$ on $P$ is same as regret of $\pi$ on $g(P)$ and hence, $\Regret(\PCoSB)\leq \Regret (\PCSB_d)$.
	Similarly, we can establish the other direction of the proposition and get $\Regret(\PCSB_d)\leq \Regret (\PCoSB)$. Thus we conclude $\Regret(\PCSB_d)= \Regret (\PCoSB)$.
\endproof

Let $\nabla_{\max}$, $\nabla_{\min}$, $\nabla_{i,\min}$, and $K^\prime$ be the same as in Section \ref{sssec:differentThetaRegretBounds}. Let $k_\star$ be the minimum number of arms in the optimal superarm and $\cM$ be the set of all feasible superarms. As it is not possible to sample $\hat\mu_i(t)$ to be precisely the true value $\mu_i$ using Beta distribution, we need to consider the $\eta$-neighborhood of $\mu_i$, and such $\eta$ term is common in the analysis of most Thompson Sampling algorithms (see \cite{NeurIPS20_perrault2020statistical,ICML18_wang2018thompson} for more details). We need the following results to prove \cref{thm:regretDiffThreshold}. 
\begin{theorem}
	\label{thm:CTSRegret}
	Let $\hat\btheta$ be allocation equivalent of $\btheta$ for instance $(\bmu,\btheta,Q)$. Then, the expected regret of \ref{alg:CSB-MT} in $T$ rounds after knowing the allocation equivalent is upper bounded by
	$$16\log_2^2(16K^\prime)\sum_{i\in [K]} \frac{ \log \left(2^{K^\prime} |\cM|T\right)}{\nabla_{i,\min}} + \nabla_{\max}(K+1) + \frac{4K (K^\prime)^2\nabla_{\max}}{\left(\nabla_{\min}-2(k_\star^2+1)\eta\right)^2}  + \nabla_{\max} \frac{C}{\eta^2}\left(\frac{C^\prime}{\eta^4}\right)^{k_\star},$$
	 where $C$, $C^\prime$ are two universal constants, and $\eta \in (0, 1)$ is such that $\nabla_{\min}-2(k_\star^2+1)\eta > 0$. Further, the expected regret of \ref{alg:CSB-MT} in $T$ rounds  is also upper bounded by $O\left(\sum\limits_{i \in [K]}\frac{\log^2(K^\prime)\log T}{\nabla_{i,\min}} \right)$.
\end{theorem}
\proof{Proof.}
	Once the allocation equivalent of $\btheta$ is known, the CSB problem is equivalent to a Combinatorial Semi-Bandit problem (from \cref{prop:MultiThetaEquivalence}). Now the proof of \cref{thm:CTSRegret} follows by verifying Assumptions $1-3$ in \cite{NeurIPS20_perrault2020statistical} for the Combinatorial Semi-Bandit problem and applying their regret bound. Assumption $1$ states that the agent has access to an oracle that can compute the optimal superarm. Whereas, Assumption $3$ states that the losses of arms are bounded and mutually independent. It is clear that both of these assumptions hold for our setting. We next proceed to verify Assumption $2$. For fix allocation $\ba\in \A$, the mean loss incurred from loss vector $\bmu$ is given by $l(M,\bmu)=\sum_{i \in M}\bmu_i$ where $M=\left\{i:a_i < \hat\theta_i\right\}$. For any two loss vectors $\bmu$ and $\bmu^\prime$, we have
	\begin{align*}
	l(M, \bmu)-l(M, \bmu^\prime)&=\sum_{i \in M}(\mu_i - \mu_i^\prime) \\
	&= \sum_{i=1}^K  \one{a_i< \hat\theta_i}\left (\mu_i -\mu_i^\prime \right) \hspace{5mm} \text{$\Bigg($as $\sum_{i \in M}\mu_i= \sum_{i=1}^K \mu_i \one{a_i< \hat\theta_i}\Bigg)$}\\
	&\leq    \sum_{i=1}^K  \left (\mu_i -\mu_i^\prime \right)\\
	&\leq    \sum_{i=1}^K   |\mu_i -\mu_i^\prime | \\
	&= B\parallel \bmu- \bmu^\prime \parallel_1
	\end{align*}
	where $B=1$. After knowing the allocation equivalent, the allocation to each arm remains the same in each round ($\hat{\theta}_i$ is given to each arm $i \in [K]\setminus A_t$). Thus we are solving a Combinatorial Semi-Bandit with parameter $B=1$. By using Theorem $1$ in \cite{NeurIPS20_perrault2020statistical}, we get the desired bounds.
\endproof

\begin{theorem}
	\label{thm:regretDiffThresholdHighConf}
	With probability at least $1-\delta$, the expected cumulative regret of \ref{alg:CSB-MT} is upper bounded as 
	\begin{align*}
		\EE{\Regret_T} \le \frac{\log( K \log_2\left(\ceil{1 +\frac{Q}{\gamma}}\right)/\delta)} {\log(1/(1-\epsilon))} &\left[\sum_{i \in A_{\Theta_n}} {\log_2 \left(\ceil{1 + \frac{Q}{\gamma}}\right)} + K{\log_2 (n +  1)}\right] \nabla_{\max} +\\
		&\qquad O\left(\sum\limits_{i \in [K]}\frac{\log^2(K^\prime)\log T}{\nabla_{i,\min}} \right).
	\end{align*}
\end{theorem}
\proof{Proof.}
	The first term of expected regret is due to the estimation of allocation equivalent. It takes $T_{\theta_n}$ rounds to complete, and $\nabla_{\max}$ is the maximum regret that can be incurred in any round. Then the maximum regret due to threshold estimation is bounded by $T_{\theta_n}\nabla_{\max}$ (replace $T_{\theta_n}$ by its value). The remaining term in regret corresponds to the expected regret incurred after knowing the allocation equivalent, that is upper bounded by \cref{thm:CTSRegret}.  
\endproof

\noindent
Let $W_\delta$  be the same as in Section \ref{sssec:differentThetaRegretBounds}. We are now ready to give the proof of \cref{thm:regretDiffThreshold}.
\regretDiffThreshold*
\proof{Proof.}
	The regret bound follows from Theorem \ref{thm:regretDiffThresholdHighConf} by setting $\delta=T^{-(\log T)^{-\alpha}}T$ and unconditioning the expected regret incurred after knowing the allocation equivalent in \ref{alg:CSB-MT}.
\endproof

\subsection*{Missing proofs from  \cref{sec:unknown} }

\sameThetaEstRounds*
\proof{Proof.}
    Let $X_1, X_2, \ldots, X_P$ be the independent Bernoulli random variables where $X_i$ has mean $\mu_i$.  The samples from all random variables are observed at the same time. Let $R_W$ is a random variable that counts the number of rounds needed to observe a sample of `$1$' for any of $\{X_i\}_{i \in [P]}$. First, we compute $\Prob{R_W = w}$, i.e.,
    \begin{equation*}
        \Prob{R_W = w} = \Pi_{i \in [P]}(1-\mu_i)^{w-1} \left(1-\Pi_{i \in [P]}(1-\mu_i)\right).
    \end{equation*}
    The previous results follows from the fact that there a sample of `$1$' is not observed for any of $\{X_i\}_{i \in [P]}$ in the first $w-1$ rounds and a sample of `$1$' is observed for at least one of the random variable in the $w^{th}$ round. The expectation of $R_W$ is given as follows:
    \begin{align*}
        \EE{R_W} &= \sum_{w=1}^{\infty} w \Prob{R_W = w}  \\
        &= \sum_{w=1}^{\infty} w \Pi_{i \in [P]}(1-\mu_i)^{w-1} \left(1-\Pi_{i \in [P]}(1-\mu_i)\right)\\
        &= \left(1-\Pi_{i \in [P]}(1-\mu_i)\right)\sum_{w=1}^{\infty} w \Pi_{i \in [P]}(1-\mu_i)^{w-1}. 
        \intertext{Let $\bar{p} = 1-\Pi_{i \in [P]}(1-\mu_i)$, we have}
        \implies \EE{R_W}  &= \bar{p} \sum_{w=1}^{\infty} w (1 - \bar{p})^{w-1} \\
        &= \bar{p}\left[ \frac{d}{d\bar{p}}\sum_{w=1}^{\infty} -(1-\bar{p})^{w}\right] \\ 
        &= \bar{p}\left[ \frac{d}{d\bar{p}}\left(\frac{-1}{\bar{p}} \right) \right] \\
        &= \bar{p}\left(\frac{1}{\bar{p}^2} \right) = \frac{1}{\bar{p}} \\
        &= \frac{1}{1- \Pi_{i \in [P]}(1-\mu_i)}.
    \end{align*}
    
    \ref{alg:CSB-JS} starts equal resources to all $L=K$ arm. When a loss is observed for any of the arms, it implies that current resource allocation is a underestimate of threshold and then resources are equally allocated among $L=K-1$ arms. Let $T_\theta(L)$ denote the number of the rounds needed to observe a loss when $L$ arms are allocated resources. By taking top $L$ arms in each round, the upper bound on expected value of $T_\theta(L)$ is given as:
    \begin{equation*}
        \EE{T_\theta(L)} \leq  \frac{1}{1 - \Pi_{i \in [K]/[K-L]}(1-\mu_i)}.
    \end{equation*}
    
    Note that $M$ is the number of arms in the optimal allocation. Consider all wrong values of $L \in \{M+1, M+2, \ldots, K-1, K \}$, the upper bound on expected number of rounds needed to reach to correct allocation, i.e., $(Q/M)$ is given as follows:
    \begin{align*}
        \EE{T_{\theta_s}} &= \sum_{L=M+1}^{K} \EE{T_\theta(L)} \\
        \implies \EE{T_{\theta_s}} &\le \sum_{L=M+1}^{K} \frac{1}{1 - \Pi_{i \in [K]/[K-L]}(1-\mu_i)}.
    \end{align*}
\endproof

\noindent
Now, we need the following results to prove the regret bound of \ref{alg:CSB-JS}.
\regretJointSameThreshold*
\proof{Proof.}
    The regret of \ref{alg:CSB-JS} can be divided into two parts: regret before knowing allocation equivalent and after knowing it. The first part of regret bound is the expected regret incurred while estimating allocation equivalent, which is $\EE{T_{\theta_s}}\nabla_{\max}$. The second part of regret is due to the MP-MAB algorithm (MP-TS) and is given by \cref{thm:MPRegret}.
\endproof

\diffThetaEstRounds*
\proof{Proof.}
    The expected number of rounds needed to observe a loss from an under-allocated arm with non-zero mean loss are $1/ \mu_i$ (by Geometric distribution). When a loss is observed for an arm, \ref{alg:CSB-JD} increments resources by $\gamma$ amount for that arm. In worse case, $\theta_i$ or more resources are allocated only after $\floor{\theta_i/\gamma}$ number of increments in resource allocation for the arm $i$. Therefore, the expected number of rounds needed to estimate $\hat{\theta}_i \in [\theta_i, \ceil{\theta_i/\gamma}\gamma]$ are $\floor{\theta_i/\gamma}(1/\mu_i)$.
    
    Let consider the worst case where only one threshold is estimated at a time. Then the maximum expected rounds needed to estimate all thresholds are $\sum_{i \in [K]:\mu_i \ne 0} \floor{{\theta_i}/{\gamma}} \left( {1}/{\mu_i} \right)$. With this argument, the proof is complete.
\endproof

\paragraph{Remark:} \ref{alg:CSB-JD} can estimate thresholds of multiple arms by starting with the same allocation of resources to all arms. Hence the number of rounds needed for finding allocation equivalent might be very small in practice than given in \cref{lem:diffThetaEstRounds} where the worst case is considered.

\regretJointDiffThreshold*
\proof{Proof.}
    Similar to \ref{alg:CSB-JS}, the regret of \ref{alg:CSB-JD} can also be divided into two parts: regret before knowing allocation equivalent and after knowing it. We get the first part of expected regret by using the upper bound on the expected number of rounds needed to find allocation equivalent from \cref{lem:diffThetaEstRounds} and the fact that $\Delta_{\max}$ is the maximum regret that can be incurred in any round. Once an allocation equivalent threshold is found, by exploiting equivalence with combinatorial semi-bandit, the second part of the expected regret is due to using a combinatorial semi-bandit algorithm (CTS-BETA) and is given by \cref{thm:CTSRegret}.
\endproof

%% file: num.tex

In this section, we study the application of CSB for the Stochastic {\em Network Utility Maximization} problem. Network Utility Maximization (NUM) is an approach for resource allocation among multiple agents such that the total utility of all the agents (network utility) is maximized. In its simplest form, NUM solves the following optimization problem:
\begin{align*}
\underset{\ba}{\text{maximize}} & \sum_{i=1}^{K}U_i(a_i)\\
 &\mbox{subject to}  \hspace{4mm}\sum_{i=1}^K a_i \leq Q
\end{align*}
where $U_i(\cdot)$ denotes the utility of agent $i$, variable $\ba=(a_1,a_2,\ldots, a_K) \in \R_+^K$ denote the allocated resource vector, and $Q \in \R_+$ is  amount of resource available. Utilities define the agents' satisfaction level, which depends on the amount of resources they are allocated. A resource could be bandwidth, power, or rates they receive. Since the seminal work of \cite{ETT1997_ChargingAndRateControl}, there has been a tremendous amount of work on NUM and its extensions. NUM is used to model various resource allocation problems and improve network protocols based on its analysis. We refer the readers to \cite{JSAC2006_TutorialOnNUM_PalomarChinag} and \cite{TAC2007_AlternateDistributed_PalomarChinag} for an informative tutorial and survey on this subject.

\begin{figure}[!ht]
	\centering 
	\includegraphics[width=0.9\linewidth]{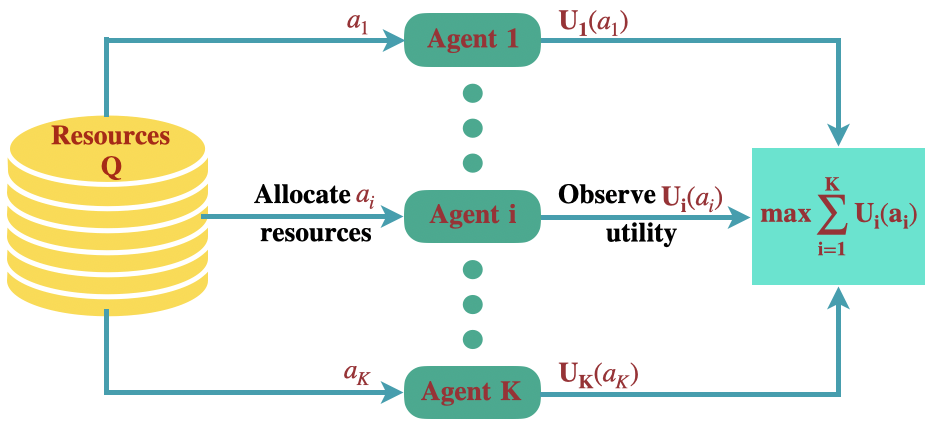} 
	\caption{Stochastic Network Utility Maximization problem where the resources are allocated among multiple agents such that the total average utility of all the agents (network utility) is maximized.}
	\label{fig:SNUM}
\end{figure}

The nature of utility functions is vital in the analysis of the NUM problem and assumed to be known or can be constructed based on the agent behavior model and operator cost model. However, agent behavior models are often difficult to quantify. Therefore, we consider the NUM problem where the utilities of the agent are unknown and stochastic. The earlier NUM problems considered deterministic settings. Significant progress has been made to extend the NUM setup to consider the stochastic nature of the network and agent behavior \citep{ETT2008_StochasticNUM_YiChiang}. For both the static and stochastic networks, the works in the literature often assume that the utility functions are smooth concave functions and apply Karush-Kuhn-Tucker conditions to find the optimal allocation. However, if the utility functions are unknown, these methods are useful only once the utilities are learned. Many of the NUM variants with full knowledge of utilities aim to find an optimal policy that meets several constraints like stability, fairness, and resource \citep{INFOCOM2010_DelayBasedNUM_Neely,WiOpt2017_DRUM_EryilmazKoprulu,WiOpt2018_NUMHetrogeneous_SinhaModiano}. In this work, we only focus on resource constraint due to limited divisible resource (bandwidth, power, rate). \cref{fig:SNUM} depicts the Stochastic Network Utility Maximization problem.

Since learning an arbitrary utility function is not always feasible,  we assume the utilities belong to a class of `threshold' type functions. Specifically, we assume that each agent's utility is stochastic with some positive mean only when it is allocated a certain minimum resource.  We refer to the minimum resource required by an agent as its `threshold' and the mean utility it receives when it is allocated resource above the threshold as its `mean reward.' Thus the expected utility of each agent is defined by two parameters -- a threshold and a mean reward.  Such threshold type utilities correspond to hard resource requirements. For example, an agent can transmit and obtain a positive rate  (reward) only if its power or bandwidth allocation is above a certain amount. In each round, the operator allocates a resource to each agent and observes the utilities the agent obtains. The goal of the operator is to allocate resources such that the expected network utility is maximized. We pose this problem as a Censored Semi-Bandits problem in the reward maximization setting, where the operator corresponds to a learner, agents to arms, and utilities to rewards. The learner's goal is to learn a policy that minimizes the difference between the best achievable expected network utility with full knowledge of the agent utilities and that obtained by the learner under the same resource constraint with the estimated utilities of agents.

\subsection{CSB setup for Reward Maximization}
The CSB problems that are considered in \cref{sec:problemSetting} works only in the loss setting. Now we extend the CSB setup to reward maximization setting, where the optimal allocation can be computed as follows:
\begin{equation*}
	\label{equ:networkUtility}
	\ba^\star  \in \argmax _{\ba \in \A} \sum_{i=1}^K\mu_i \one{a_i\ge \theta_i}.
\end{equation*}

The interaction between a learner and the environment that governs rewards for the arms is as follows: In the round $t$, the environment generates a reward vector $(X_{t,1}, X_{t,2},\ldots, X_{t,K}) \in \{0,1\}^K$, where $X_{t,i}$ denotes the true reward for arm $i$ in round $t$.  The sequence $(X_{t,i})_{t\geq 1}$ is generated IID with the common mean $\EE{X_{t,i}}=\mu_i$ for each $i \in [K]$. The learner selects a feasible allocation $\ba_t=\{a_{t,i}: i\in [K]\}$ and observes reward vector $X_t^\prime=\{X^\prime_{t,i}: i\in [K]\}$, where $X_{t,i}^\prime=X_{t,i}\one{a_{t,i}\ge \theta_i}$ and collects reward $r_t(\ba_t)=\sum_{i \in [K]}X_{t,i}^\prime$. A policy of the learner is to select a feasible allocation in each round based on the observed reward such that the cumulative reward is maximized. The performance of a policy that makes allocation $\{\ba_t\}_{t\geq1}$ in round $t$ is measured in terms of expected regret for $T$ rounds given by

\begin{equation*}
	\mathbb{E}[\Regret_T] = T\sum_{i=1}^K\mu_i \one{x^\star_i\ge \theta_i} - \EE{ \sum_{t=1}^T\sum_{i=1}^K Y_{t,i} \one{x_{t,i}\ge \theta_i}}.
\end{equation*}

A good policy must have sub-linear regret, i.e., $\EE{\Regret_T}/T \rightarrow 0$ as $T \rightarrow \infty$. Next, we define the notion of treating a pair of thresholds for the given reward vector and resource to be `equivalent.'
\begin{definition}[Allocation Equivalent]
	For fixed reward vector $\bmu$ and amount of resource $Q$, two threshold vectors $\btheta$ and $\hat{\btheta}$ are {\em allocation equivalent} if the following holds:
	\begin{equation*}
		\max_{\ba \in \A} \sum_{i=1}^K\mu_i \one{a_i\ge \theta_i}  = \max_{\ba \in \A} \sum_{i=1}^K\mu_i \one{a_i\ge \hat{\theta}_i}.
	\end{equation*}
\end{definition}

\subsection{Algorithms for Network Utility Maximization}
We first focus on the special case of the network utility maximization problem where $\theta_i=\theta_s$ for all $i \in [K]$. We develop an algorithm named Network Utility Maximization with the Same Threshold (\ref{alg:ONUM-ST}). This algorithm is adapted from \ref{alg:CSB-ST} for the reward maximization setup. There are two major differences: 1) the feedback (reward) is only observed when the resource allocation is more than a certain threshold, and 2) a sample for the mean reward estimate is drawn from the beta distribution. Similarly, we develop an algorithm named Network Utility Maximization with the Multiple Threshold (\ref{alg:ONUM-DT}), which is adapted from \ref{alg:CSB-MT} to the reward maximization setup.

\begin{algorithm}[!ht] 
	\renewcommand{\thealgorithm}{\bf NUM-SK}
	\floatname{algorithm}{}
	\caption{Algorithm for NUM problem having Same Threshold with Known Horizon and $\epsilon$}
	\label{alg:ONUM-ST}
	\begin{algorithmic}[1]
		\State \textbf{Input:} $\delta, \epsilon$
		\State Set $W_\delta = {\log(\log_2(K)/\delta)}/({\log(1/(1-\epsilon))})$ and $\forall i \in [K]: S_i=1, F_i=1, Z_i=0$ 
		\State Initialize $\Theta$ as given in Lemma \ref{lem:thetaSet}, $C=0, l=1, u = K, j = \floor{(l+u)/2}$ 
		\For{$t=1,2,\ldots,$}
		\State Set $\hat{\theta}_s = \Theta[j]$ and $\forall i \in [K]: \hat{\mu}_{t,i} \leftarrow \beta(S_i, F_i)$
		\State $A_t \leftarrow$ set of top-$({Q}/{\hat{\theta}_s})$ arms with the largest values of $\hat{\mu}_{t,i}$ 
		\State $\forall i \in A_t:$ allocate $\hat{\theta}_s$ resource and observe $X_{t,i}$
		\If{$j \ne u$} 
		\If{$X_{t,a} = 1$ for any $a \in A_t$} 
		\State Set $u=j, ~j = \floor{(l+u)/2}, C=0$
		\State $\forall i \in A_t$: set $S_i = S_i+X_{t,i}, F_i = F_i+1-X_{t,i} + Z_i, Z_i =0$
		\State $\forall i \in [K]\setminus A_t: F_i = F_i+ Z_i, Z_i =0$
		\Else
		\State Set $C = C + 1$ and $\forall i \in A_t: Z_i = Z_i + 1$
		\State If $C = W_\delta$ then set $l=j+1, j = \floor{(l+u)/2}$, $C=0, \forall i \in [K]: Z_i =0$
		\EndIf
		\Else
		\State $\forall i \in A_t: S_i = S_i + X_{t,i}, F_i = F_i + 1 - X_{t,i}$
		\EndIf
		\EndFor
	\end{algorithmic}
\end{algorithm}

Next, we will give the regret upper bounds for \ref{alg:ONUM-ST} and \ref{alg:ONUM-DT}. For simplicity of discussion, we assume that arms are indexed according to their decreasing mean rewards, i.e., $\mu_1 \geq \mu_2, \ldots, \geq \mu_K$, but the algorithms are not aware of this ordering. We refer to the first $M$ arms as \emph{top-}$M$ arms. For a instance $(\bmu,\btheta, C)$ and any feasible allocation $\ba \in \A$, we define the sub-optimality gap as $\Delta_a = \sum_{i=1}^K\mu_i\big(\one{a_i^\star \ge \theta_i} - \one{a_i \ge \theta_i}\big)$. The maximum and minimum regret incurred in a round is $\nabla_{\max} = \max\limits_{\ba \in \A } \nabla_{\ba}$ and $\nabla_{\min} = \min\limits_{\ba \in \A } \nabla_{\ba}$, respectively. Now we will give regret bound of \ref{alg:ONUM-ST}.

\begin{algorithm}[!ht]
	\small 
	\renewcommand{\thealgorithm}{\bf NUM-MK} 
	\floatname{algorithm}{}
	\caption{Algorithm for NUM problem having Multiple Threshold with Known Horizon and $\epsilon$}
	\label{alg:ONUM-DT}
	\begin{algorithmic}[1]
		\State \textbf{Input:} $n, \delta, \epsilon, \gamma$
		\State Initialize: $\forall i \in [K]: S_i = 1, F_i = 1, Z_i =0, \theta_{l,i} = 0, \theta_{u,i} = Q, \theta_{g,i} = 0,  \hat\theta_{i} = Q/2,$ 
		\State Set $\Theta_n = \emptyset,W_\delta = \log (K\log_2(\lceil 1 + Q/\gamma\rceil)/\delta)/\log(1/(1-\epsilon)),$ if $n<K$ then $n_i =1$ else $n_i=0$ 
		\For{$t=1,2, \ldots,$}
		\State $\forall i \in [K]: \hat{\mu}_{t,i} \leftarrow \text{Beta}(S_i, F_i)$
		\If{$\theta_{g,j} = 0$ for any $j \in [K]$}
		\If{$n < K$}
		\While{$\theta_{g,n_ i} = 1$} 
			\State Add $\hat\theta_{n_i}$ to $\Theta_n$ and set $n_i = n_i + 1$.  Sort $\Theta_n$ in increasing order
		\EndWhile
		\If{there exists no $j \in [|\Theta_n|]$ such that $\theta_{l,n_i} < \Theta_n[j] \le \theta_{u,n_i}$ or $\Theta_n = \emptyset$}
		\State  Set $\hat\theta_{n_i}=(\theta_{l,i}+\theta_{u,i})/2$ 
		\Else
		\State Set $l = \min\{k: \Theta_n[k] > \theta_{l,n_i}\}, u = \max\{k: \Theta_n[k] \le \theta_{u,n_i}\},$ and $j = \floor{(l+u)/2}$
		\State If $\Theta_n[j]=\theta_{u,n_i}$ then set $\hat\theta_{n_i} = \Theta_n[j]-\gamma$ else  $\hat\theta_{n_i} = \Theta_n[j]$
		\EndIf
		\EndIf
		
		\State $\forall i \in [K]\setminus \{n_i\}$: update $\hat\theta_{i}$ using Eq. \eqref{equ:updateTheta}. Allocate $\hat\theta_{i}$ resource to arm $i$ and observe $X_{t,i}$
		\For{$i = \{1,2,\ldots, K\}$}
		\If{$\theta_{g,i} = 0$ and $\hat\theta_{i}> \theta_{l,i}$}
		\State If $X_{t,i}=1$ then set $\theta_{u,i} = \hat\theta_{i}, S_i = S_i + 1, F_i = F_i + Z_i, Z_i =0$ else  $Z_i = Z_i + 1$
		\State If {$Z_i= W_\delta$} then set $\theta_{l,i} = \hat\theta_{i}, Z_i=0 $
		\State If $\theta_{u,i} - \theta_{l,i} \le \gamma$ then set $\theta_{g,i}=1$ and $\hat\theta_i = \theta_{u,i}$
		\ElsIf{$\hat\theta_{i} \ge \theta_{u,i}$ or $\big\{ \theta_{g,i} = 1$ and $\hat\theta_{i}\ge \hat\theta_{u,i} \big\}$}
		\State Set $S_i = S_i+X_{t,i}$ and $F_i = F_i+1-X_{t,i}$ 
		\EndIf	
		\EndFor	
		
		\Else
		\State $A_t \leftarrow$ Oracle$\big( KP(\hat\bmu_{t}, \hat\btheta, C)\big)$
		\State $\forall i \in A_t:$ allocate $\hat\theta_{i}$ resource and observe $X_{t,i}$, update $S_i = S_i+X_{t,i}$ and $F_i = F_i+1-X_{t,i}$
		\EndIf
		\EndFor
	\end{algorithmic}
\end{algorithm}

\begin{theorem}
	\label{thm:regretONUM_ST}
	Let $\mu_K\geq \epsilon>0$, $\mu_{M} > \mu_{M+1}$, $W_\delta = {\log(\log_2(K)/\delta)}/{\log(1/(1-\epsilon))}$, and $T>W_\delta\log_2{(K)}$. Then with probability at least $1-\delta$, the expected regret of \ref{alg:ONUM-ST} is upper bounded as
	\begin{align*}
		\EE{\Regret_T} &\le  W_\delta\log_2{(K)}\Delta_{\max} + O\left((\log T)^{{2}/{3}}\right) + \mbox{$\sum_{i \in [K]\setminus [M]}$} \frac{(\mu_M-\mu_i)\log {T}}{d( \mu_i,\mu_M)}.
	\end{align*}
\end{theorem}
\noindent
As we are in the reward setting, the regret bound of MP-TS can be used as it is. The remaining proof follows similar steps as the proof of \cref{thm:regretSameThreshold}.

Let $\nabla_{i, \min}$ be the minimum regret for superarms containing arm $i$ and $K^\prime$ be the maximum number of arms in any feasible resource allocation. We redefine $W_\delta = \log (K\log_2(\lceil 1 + Q/\gamma\rceil)/\delta)/\log(1/(1-\epsilon))$. Now we are ready to state the regret bound of \ref{alg:ONUM-DT}.
\begin{theorem}
	\label{thm:regretUNUM_DT}
	Let $\gamma >0$, $\mu_K\geq \epsilon>0$, and $T>KW_\delta\log_2 \left(\lceil 1+Q/\gamma\rceil \right)$. Then with probability at least $1-\delta$, the expected regret of \ref{alg:ONUM-DT} is upper bounded by 
	\begin{align*}
		\EE{\Regret_T} &\le {KW_\delta\log_2 \left(\lceil 1+Q/\gamma\rceil \right)\Delta_{\max}}  +  O\left(\sum_{i \in [K]}\frac{\log^2(K^\prime)\log T}{\nabla_{i, \min}} \right).
	\end{align*}
\end{theorem}
\noindent
Since we are in the reward setting, the regret bound of combinatorial bandits algorithm CTS-BETA can be used as it is. The remaining proof follows similar steps as the proof of \cref{thm:regretDiffThreshold}. 

\paragraph{Anytime Algorithms for reward maximization setting.}
For simplicity, consider the reward setting with a single threshold. When the allocated resource exceeds the arm's threshold, the learner may continue to observe sample values of $0$ due to the stochastic nature of reward generation. Thus, the learner needs to observe enough samples to be confident that the resource allocated is above the threshold. To decide how much is enough, the learner needs to know $T$ so that exploration and exploitation can be well balanced. However, note that this issue does not arise in the loss setting; if the learner continues to observe a sample $0$, there is no need to increase the allocation further, and the learner can continue the same resource allocation on the arm. The same argument applied if the learner has to start by allocating the higher amount of resources and keep decreasing it until it goes below the threshold.

\vspace{2mm}